\documentclass[letterpaper, 10 pt, conference]{ieeeconf}  

\IEEEoverridecommandlockouts                              
\makeatletter
\let\NAT@parse\undefined
\renewcommand{\@IEEEsectpunct}{\ \,}
\makeatother

\usepackage[dvipsnames]{xcolor}

\newcommand*\linkcolours{Red}

\usepackage{graphicx}
\usepackage{amsmath, amssymb}

\usepackage{amsthm}

\usepackage{url}
\hypersetup{
colorlinks,
linkcolor=\linkcolours,
citecolor=\linkcolours,
filecolor=\linkcolours,
urlcolor=\linkcolours}

\usepackage[ruled]{algorithm}
\usepackage{algpseudocode}

\usepackage[labelfont={bf},font=small]{caption}
\usepackage[none]{hyphenat}

\usepackage{tabularx}

\usepackage{comment}

\DeclareMathAlphabet{\mathpzc}{OT1}{pzc}{m}{it}

\DeclareMathOperator*{\argsup}{arg\,sup}

\newtheorem{theorem}{Theorem}
\newtheorem{definition}[theorem]{Definition}

\newtheorem{lemma}[theorem]{Lemma}

\usepackage{xspace}
\usepackage{dsfont} 
\newcommand{\ie}{\textrm{i.e.}}

\newcommand{\ccite}[1]{\xspace\cite{#1}}

\newcommand{\sref}{Section~\ref}
\newcommand{\appref}[1]{Appendix~\ref{#1}}
\newcommand{\aref}[1]{Algorithm~\ref{#1}}
\newcommand{\eref}[1]{eq.(\ref{#1})}
\newcommand{\fref}[1]{Figure~\ref{#1}}

\newcommand{\tref}[1]{Table~\ref{#1}}
\newcommand{\lref}[1]{Lemma~\ref{#1}}

\newcommand{\dref}[1]{Definition~\ref{#1}}
\newcommand{\thref}[1]{Theorem~\ref{#1}}

\newcounter{comment}

\newcommand{\monName}{Statistical-distance-based Non-linearity Measure\xspace}
\newcommand{\monNameAbbr}{SNM\xspace}

\newcommand{\nop}{SNM-Planner\xspace}

\newcommand{\monG}{Measure of Non-Gaussianity\xspace}
\newcommand{\monGAbbr}{MoNG\xspace}
\newcommand{\mong}{Measure of Non-Gaussianity\xspace}
\newcommand{\mongAbbr}{MoNG\xspace}

\newcommand{\pomdpTuple}{\ensuremath{\langle S, A, O, T, Z, R, \belInit, \gamma \rangle}\xspace}
\newcommand{\linPomdpTuple}{\ensuremath{\langle S, A, O, \widehat{T}, \widehat{Z}, R, \belInit, \gamma \rangle}\xspace}
\newcommand{\pomdp}{\ensuremath{P}\xspace}
\newcommand{\linPomdp}{\ensuremath{\widehat{P}}\xspace}

\newcommand{\nm}[2]{\ensuremath{\Psi(#1, #2)}\xspace}
\newcommand{\nmSym}{\ensuremath{\Psi}\xspace}

\newcommand{\nmSymT}{\ensuremath{\Psi_T}\xspace}
\newcommand{\nmSymZ}{\ensuremath{\Psi_Z}\xspace}
\newcommand{\nmT}[2]{\ensuremath{\Psi_T(#1, #2)}\xspace}
\newcommand{\nmZ}[2]{\ensuremath{\Psi_Z(#1, #2)}\xspace}

\newcommand{\J}[2]{\ensuremath{J(#1, #2)}\xspace}

\newcommand{\trans}{\ensuremath{T}\xspace}
\newcommand{\transComp}{\ensuremath{T(s, a, s')}\xspace}
\newcommand{\transGauss}{\ensuremath{T_G}\xspace}
\newcommand{\transCompGauss}{\ensuremath{T_G(s, a, s')}\xspace}
\newcommand{\obsF}{\ensuremath{Z}\xspace}
\newcommand{\obsFGauss}{\ensuremath{Z_G}\xspace}
\newcommand{\obsFComp}{\ensuremath{Z(s, a, o)}\xspace}

\newcommand{\obsFCompGauss}{\ensuremath{Z_G(s, a, o)}\xspace}
\newcommand{\linTrans}{\ensuremath{\widehat{T}}\xspace}
\newcommand{\linTransComp}{\ensuremath{\widehat{T}(s, a, s')}\xspace}
\newcommand{\linObsF}{\ensuremath{\widehat{Z}}\xspace}
\newcommand{\linObsFComp}{\ensuremath{\widehat{Z}(s, a, o)}\xspace}

\newcommand{\rewFuncComp}[2]{\ensuremath{R(#1, #2)}\xspace}

\newcommand{\optPol}{\ensuremath{\pi^*}\xspace}
\newcommand{\linOptPol}{\ensuremath{\widehat{\pi}^*}\xspace}

\newcommand{\lin}[1]{\ensuremath{\widehat{#1}}\xspace}

\newcommand{\pet}{\ensuremath{e_T}\xspace}
\newcommand{\pez}{\ensuremath{e_Z}\xspace}

\newcommand{\stSpace}{\ensuremath{S}\xspace}

\newcommand{\st}{\ensuremath{s}\xspace}

\newcommand{\stp}{\ensuremath{s'}\xspace}
\newcommand{\actSpace}{\ensuremath{A}\xspace}

\newcommand{\act}{\ensuremath{a}\xspace}
\newcommand{\obsSpace}{\ensuremath{O}\xspace}

\newcommand{\obs}{\ensuremath{o}\xspace}

\newcommand{\bel}{\ensuremath{b}\xspace}

\newcommand{\belInit}{\ensuremath{b_0}\xspace}
\newcommand{\belS}[1]{\ensuremath{b(#1)}\xspace}

\newcommand{\entr}[1]{\ensuremath{H(#1)}\xspace}

\newcommand{\alphaFunctPolComp}[2]{\ensuremath{\alpha_{#1}(#2)}\xspace}
\newcommand{\linAlphaFunctPolComp}[2]{\ensuremath{\widehat{\alpha}_{#1}(#2)}\xspace}

\newcommand{\alphaPiHatS}[1]{\ensuremath{\widehat{\alpha}_{\sigma}(#1)}\xspace}
\newcommand{\alphaPiS}[1]{\ensuremath{\alpha_{\sigma}(#1)}\xspace}

\newcommand{\intS}{\int_{s \in S}\xspace}
\newcommand{\intSDash}{\ensuremath{\int_{s' \in S}}\xspace}
\newcommand{\intO}{\ensuremath{\int_{o \in O}}\xspace}

\newcommand{\de}{\ensuremath{d}\xspace}

\title{\LARGE \bf
Non-Linearity Measure for POMDP-based Motion Planning
}

\author{Marcus Hoerger$^{1}$ and Hanna Kurniawati$^{1}$ and Alberto Elfes$^{2}$
\thanks{$^{1}$Research School of Computer Science.
Email: \{marcus.hoerger, hanna.kurniawati\}@anu.edu.au}%
\thanks{$^{2}$Robotics and Autonomous Systems Group, Data61, CSIRO.
Email: alberto.elfes@data61.csiro.au}%
}

\begin{document}

\maketitle
\thispagestyle{empty}
\pagestyle{empty}

\begin{abstract}

Motion planning under uncertainty is essential for reliable robot operation. Despite substantial advances over the past decade, the problem remains difficult for systems with complex dynamics. Most state-of-the-art methods perform search that relies on a large number of forward simulations. For systems with complex dynamics, this generally require costly numerical integrations which significantly slows down the planning process. Linearization-based methods have been proposed that can alleviate the above problem. However, it is not clear how linearization affects the quality of the generated motion strategy, and when such simplifications are admissible. We propose a non-linearity measure, called Statistical-distance-based Non-linearity Measure (SNM), that can identify where linearization is beneficial and where it should be avoided. We show that when the problem is framed as the Partially Observable Markov Decision Process, the value difference between the optimal strategy for the original model and the linearized model can be upper bounded by a function linear in SNM. Comparisons with an existing measure on various scenarios indicate that SNM is more suitable in estimating the effectiveness of linearization-based solvers. To test the applicability of SNM in motion planning, we propose a simple on-line planner that uses SNM as a heuristic to switch between a general and a linearization-based solver. Results on a car-like robot with second order dynamics and 4-DOFs and 7-DOFs torque-controlled manipulators indicate that SNM can appropriately decide if and when a linearization-based solver should be used.

\end{abstract}

\section{Introduction }
\label{section:introduction}
An autonomous robot must be able to compute reliable motion strategies, despite various errors in actuation and prediction on its effect on the robot and its environment, and despite various errors in sensors and sensing. Computing such robust strategies is computationally hard even for a 3 DOFs point robot\ccite{Can87:New},\ccite{Nat88:Complexity}. Conceptually, this problem can be  solved in a systematic and principled manner when framed as the Partially Observable Markov Decision Process (POMDP)\ccite{Kae98:Planning}. A POMDP represents the aforementioned errors as probability distribution functions and estimates the state of the system as probability distribution functions called \emph{beliefs}. It then computes the best motion strategy with respect to beliefs rather than single states, thereby accounting the fact that the actual state is never known due to the above errors. Although the concept of POMDPs was proposed in the '60s\ccite{Son71:The}, only recently that POMDPs started to become practical for robotics problems (e.g.\ccite{Hoe19:POMDP,Hor13:Interactive,Tem09:Unmanned}). This advancement is achieved by trading optimality with approximate optimality for speed and memory. But even then, in general, computing close to optimal POMDP solutions for systems with complex dynamics remains difficult.

Several general POMDP solvers ---solvers that do not restrict the type of dynamics and sensing model of the system, nor the type of distributions used to represent uncertainty--- can now compute good motion strategies on-line with a 1-10Hz update rate for a number of robotic problems\ccite{Kur13:An,Sil10:Monte,Som13:Despot,Sei15:An}. 
However, their speed degrades when the robot has complex non-linear dynamics. 
To compute a good strategy, today's POMDP solvers forward simulate  the effect of many sequences of actions from different beliefs are simulated. For problems whose dynamics have no closed-form solutions, a simulation run generally invokes many numerical integrations, and complex dynamics tend to increase the cost of each numerical integration, which in turn significantly increases the total planning cost of these methods. Of course, this cost will increase even more for problems that require more or longer simulation runs, such as in problems with long planning horizons.

Many linearized-based POMDP solvers have been proposed\ccite{Sun15:High,Agh13:Firm,Ber10:LQGMP,Ber12:LQG,Pre10:The}. They  rely on many forward simulations from different beliefs too, but use a linearized model of the dynamics and sensing for simulation. Together with linearization, many of these methods assume that beliefs are Gaussian distributions. This assumption improves the speed of simulation further, because the subsequent belief after an action is performed and an observation is perceived can be computed in closed-form. In contrast, the aforementioned general solvers typically represent beliefs as sets of particles and estimate subsequent beliefs using particle filters. Particle filters are particularly expensive when particle trajectories have to be simulated and each simulation run is costly, as is the case for motion-planning of systems with complex dynamics. As a result, the linearization-based planners require less time to estimate the effect of performing a sequence of actions from a belief, and therefore can \emph{potentially} find a good strategy faster than the general method. However, it is known that linearization in control and estimation performs well only when the system's non-linearity is ``weak"\ccite{Li12:Measure}. The question is, what constitute ``weak" non-linearity in motion planning under uncertainty? Where will it be useful and where will it be damaging to use linearization (and Gaussian) simplifications? 

This paper extends our previous work\ccite{Hoe16:Linearization} towards answering the aforementioned questions. Specifically, we propose a measure of non-linearity for stochastic systems, called \emph{\monName (\monNameAbbr)}, to help identify the suitability of linearization in a given problem of motion planning under uncertainty. \monNameAbbr is based on the total variation distance between the original dynamics and sensing models, and their corresponding linearized models. It is general enough to be applied to any type of motion and sensing errors, and any linearization technique, regardless of the type of approximation of the true beliefs (e.g., with and without Gaussian simplification). We show that the difference between the value of the optimal strategy generated if we plan using the original model and if we plan using the linearized model, can be upper bounded by a function linear in \monNameAbbr. Furthermore, our experimental results indicate that compared to recent state-of-the-art methods of non-linearity measures for stochastic systems, \monNameAbbr is more sensitive to the effect that obstacles have on the effectiveness of linearization, which is critical for motion planning.

To further test the applicability of \monNameAbbr in motion planning, we develop a simple on-line planner that uses a local estimate of \monNameAbbr to automatically switch between a general planner\ccite{Kur13:An} that uses the original POMDP model and a linearization-based planner (adapted from\ccite{Sun15:High}) that uses the linearized model. Experimental results on a car-like robot with acceleration control, and a 4-DOFs and 6-DOFs manipulators with torque control indicate that this simple planner can appropriately decide if and when linearization should be used and therefore computes better strategies faster than each of the component planner.

\section{Background and Related Work}\label{sec:relWork}
\subsection{Background}
In this paper, we consider motion planning problems, in which a robot must move from a given initial state to a state in the goal region while avoiding obstacles. The robot operates inside deterministic, bounded, and perfectly known 2D or 3D environments populated by static obstacles.

The robot's transition and observation models are uncertain and defined as follows. 
Let $\stSpace \subset \mathbb{R}^n$ be the bounded n-dimensional state space, $A \subset \mathbb{R}^d$ the bounded $d$-dimensional control space and $\obsSpace \subset \mathbb{R}^l$ the bounded $l$-dimensional observation space of the robot. 
The state of the robot evolves according to a discrete-time non-linear function, which we model in the general form $\st_{t+1} = f(\st_t, \act_t, v_t)$ where $\st_t \in \stSpace$ is the state of the robot at time $t$, $\act_t \in \actSpace$ is the control input at time $t$, and $v_t \in \mathbb{R}^d$ is a random transition error. At each time step $t$, the robot perceives imperfect information regarding its current state according to a non-linear stochastic function of the form $\obs_t = h(\st_t, w_t)$, where $\obs_t \in \obsSpace$ is the observation at time $t$ and $w_t \in \mathbb{R}^d$ is a random observation error.

This class of motion planning problems under uncertainty can naturally be formulated as a Partially Observable Markov Decision Process (POMDP). Formally, a POMDP is a tuple \pomdpTuple, where \stSpace, \actSpace and \obsSpace are the state, action, and observation spaces of the robot. $T$ is a conditional probability function $T(\st, \act, \stp) = p(\stp \,|\, \st, \act)$ (where $\st, \stp \in \stSpace$ and $\act \in \actSpace$) that models the uncertainty in the effect of performing actions, while $Z(\st, \act, \obs) = p(\obs | \st, \act)$ (where $\obs\in\obsSpace$) is a conditional probability function that models the uncertainty in perceiving observations. $R(\st, \act)$ is a reward function, which encodes the planning objective. \belInit is the initial belief, capturing the uncertainty in the robot's initial state and $\gamma \in (0, 1)$ is a discount factor. 

At each time-step, a POMDP agent is at a state $s \in \stSpace$, takes an action $a \in \actSpace$, perceives an observation $o \in \obsSpace$, receives a reward based on the reward function $R(s, a)$, and moves to the next state. Now due to uncertainty in the results of action and sensing, the agent never knows its exact state and therefore, estimates its state as a probability distribution, called belief. The solution to the POMDP problem is an optimal policy (denoted as \optPol), which is a mapping $\optPol: \mathbb{B} \rightarrow \actSpace$ from beliefs ($\mathbb{B}$ denotes the set of all beliefs, which is called the belief space) to actions that maximizes the expected total reward the robot receives, i.e.
\begin{align}
&V^*(\belInit) = \nonumber\\ &\max_{a \in \actSpace} \left(R(b, a) + \gamma \int_{o \in \obsSpace} p(o | b, a) V^*(\tau(b, a, o)) \, \de\obs\right)
\end{align}
where $\tau(b, a, o)$ computes the updated belief estimate after the robot performs action $a \in \actSpace$ and perceived $o \in \obsSpace$ from belief $b$, and is defined as:  
\begin{align}\label{e:belTrans}
b'(s') &= \tau(b, a, o)(s') \nonumber \\&= \eta \, Z(s', a, o) \int_{s \in \stSpace} T(s, a, s') b(s) ds
\end{align}

For the motion planning problems considered in this work, we define the spaces $S$, $A$, and $O$ to be the same as those of the robotic system (for simplicity, we use the same notation). The transition $T$ represents the dynamics model $f$, while $Z$ represents the sensing model $h$. The reward function represents the task' objective, for example, high reward  for goal states and low negative reward for states that cause the robot to collide with the obstacles. The initial belief \belInit represents uncertainty on the starting state of the robot. 

\subsection{Related Work on Non-Linearity Measures}
Linearization is a common practice in solving non-linear control and estimation problems. It is known that linearization performs well only when the system's non-linearity is ``weak"\ccite{Li12:Measure}. To  identify the effectiveness of linearization in solving non-linear problems, a number of non-linearity measure have been proposed in the control and information fusion community. 

Many of these measures (e.g.\ccite{Bat80:Relative,Bea60:Confidence,Ema93:A}) have been designed for deterministic systems. For instance,\ccite{Bat80:Relative} proposed a measure derived from the curvature of the non-linear function. The work in\ccite{Bea60:Confidence,Ema93:A} computes a measure based  on the distance between the non-linear function and its nearest linearization. A brief survey of non-linearity measures for deterministic systems is available in\ccite{Li12:Measure}.

Non-linearity measures for stochastic systems has been proposed. For instance,\ccite{Li12:Measure}  extends the measures in\ccite{Bea60:Confidence,Ema93:A} to be based on the average distance between the non-linear function that models the motion and sensing of the system, and the set of all possible linearizations of the function. 

Another example is\ccite{Dun13:Nonlinearity} that  proposes a measures which is based on the distance between distribution over states and its Gaussian approximation, called \mong (\mongAbbr), rather than based on the non-linear function itself. Assuming a passive stochastic systems, this measures computes the negentropy between a transformed belief and its Gaussian approximation. The results indicate that this measure is more suitable to measure the non-linearity of stochastic systems, as it takes into account the effect that non-linear transformations have on the shape of the transformed beliefs. This advancement is encouraging and we will use \mongAbbr as a comparator of \monNameAbbr. However, for this purpose, \mongAbbr must be modified since we consider non-passive problems in work. The exact modifications we made can be found in \sref{ssec:MHFR}. 

Despite the various non-linearity measures that have been proposed, most are not designed to take the effect of obstacles to the non-linearity of the system into account. Except for \mongAbbr, all of the aforementioned non-linearity measures will have difficulties in reflecting these effects, even when they are embedded in the motion and sensing models. For instance, curvature-based measures requires the non-linear function to be twice continuously differentiable, but the presence of obstacles is very likely to break the differentiability of the motion model. Furthermore, the effect of obstacles is likely to violate the additive Gaussian error,  required for instance by\ccite{Li12:Measure}. Although \mongAbbr can potentially take the effect of obstacles into account, it is not designed to. In the presence of obstacles, beliefs have support only in the valid region of the state space, and therefore computing the difference between beliefs and their Gaussian approximations is likely to underestimate the effect of obstacles. 

\monNameAbbr is designed to address these issues. Instead of building upon existing non-linearity measures, \monNameAbbr adopts approaches commonly used for sensitivity analysis\ccite{Mas12:Loss,Mul97:Does} of Markov Decision Processes (MDP) ---a special class of POMDP where the observation model is perfect, and therefore the system is fully observable. These approaches use statistical distance measures between the original transition dynamics and their perturbed versions. Linearized dynamics can be viewed as a special case of perturbed dynamics, and hence this statistical distance measure can be applied as a non-linearity measure, too. We do need to extend these analysis, as they are generally defined for discrete state space and are defined with respect to only the transition models (MDP assumes the state of the system is fully observable). Nevertheless, such extensions are feasible and the generality of this measure could help identifying the effectiveness of linearization in motion planning under uncertainty problems.

\section{SNM}\label{sec:SNM}
Intuitively, our proposed measure \monNameAbbr is based on the total variation distance between the effect of performing an action and perceiving an observation under the true dynamics and sensing model, and the effect under the linearized dynamic and sensing model. The total variation distance $D_{TV}$ between two probability measures $\mu$ and $\nu$ over a measurable space $\Omega$ is defined as $D_{TV}(\mu, \nu) = \sup_{E \in \Omega} \left |\mu(E) - \nu(E) \right |$. An alternative expression of $D_{TV}$ which we use throughout the paper is the functional form $D_{TV}(\mu, \nu) = \frac{1}{2}\sup_{\left |f \right | \leq 1}\left |\int f\de\mu - \int f\de\nu \right |$.
Formally, \monNameAbbr is defined as:
\begin{definition}
\label{def:mon}
Let $\pomdp = \pomdpTuple$ be the POMDP model of the system and $\linPomdp = \linPomdpTuple$ be a linearization of \pomdp, where \linTrans is a linearization of the transition function \trans and \linObsF is a linearization of the observation function \obsF of \pomdp, while all other components of \pomdp and \linPomdp are the same. Then, the \monNameAbbr (denoted as \nmSym) between \pomdp and \linPomdp is $\nm{\pomdp}{\linPomdp} = \nmT{\pomdp}{\linPomdp} + \nmZ{\pomdp}{\linPomdp}$, where 
\begin{align}
\nmT{\pomdp}{\linPomdp} &= \sup_{s \in S, a \in A} D_{TV}(\transComp, \linTransComp)  \\
\nmZ{\pomdp}{\linPomdp} &= \sup_{s \in S, a \in A} D_{TV}(\obsFComp, \linObsFComp)
\end{align}
\end{definition}
Note that \monNameAbbr can be applied as both a global and a local measure. In the latter case, the supremum over the state $s$ can be restricted to a subset of \stSpace, rather than the entire state space. Furthermore, \monNameAbbr is general enough for any approximation to the true dynamics and sensing model, which means that it can be applied to any type of linearization and belief approximation techniques, including those that assume and those that do not assume Gaussian belief simplifications. 

We want to use the measure \nm{\pomdp}{\linPomdp} to bound the difference between the expected total reward received if the system were to run the optimal policy of the true model \pomdp and if it were to run the optimal policy of the linearized model \linPomdp. Note that since our interest is in the actual reward received, the values of these policies are evaluated with respect to the original model \pomdp (we assume \pomdp is a faithful model of the system). More precisely, we want to show that:
\begin{theorem}\label{th:valUpperBound}
If \optPol denotes the optimal policy for \pomdp and \linOptPol denotes the optimal policy for \linPomdp, then for any $\bel\in\mathbb{B}$, 
\begin{align}
&V_{\optPol}(\bel) - V_{\linOptPol}(\bel) \leq 4\gamma\frac{R_{max}}{(1-\gamma)^2} \nm{\pomdp}{\linPomdp} \nonumber
\end{align}
where \newline
$V_{\pi}(b) = R(b, \pi(b)) + \gamma \int_{o \in O}Z(b, a, o)V_{\pi}(\tau(b, a, o)) \de\obs$ for any policy $\pi$ and $\tau(b, a, o)$ is the belief transition function as defined in \eref{e:belTrans}
\end{theorem}

To proof \thref{th:valUpperBound}, we first assume, without loss of generality, that a policy $\pi$ for a belief \bel is represented by a conditional plan $\sigma\in\Gamma$, where $\Gamma$ is the set of all conditional plans. $\sigma$ can be specified by a pair $\left \langle \act, \nu \right \rangle$, where $\act\in\actSpace$ is the action of $\sigma$ and $\nu: \obsSpace \rightarrow \Gamma$ is an observation strategy which maps an observation to a conditional plan $\sigma'\in\Gamma$.

Every $\sigma$ corresponds to an $\alpha$-function $\alpha_{\sigma}: \stSpace \rightarrow \mathbb{R}$ which specifies the expected total discounted reward the robot receives when executing $\sigma$ starting from $\st\in\stSpace$, i.e.
\begin{align}\label{eq:alpha_s}
&\alpha_{\sigma}(s) = \rewFuncComp{\st}{\act}\nonumber\\ &+ \gamma \intSDash \intO T(\st, \act, \stp) Z(\stp, \act, \obs)\alpha_{\nu(\obs)}(\stp) \de\obs \de\stp
\end{align} 

where $\act\in\actSpace$ is the action of $\sigma$ and $\alpha_{\nu(\obs)}$ is the $\alpha$-function corresponding to conditional plan $\nu(\obs)$.

For a given belief \bel, the value of the policy $\pi$ represented by the conditional plan $\sigma$ is then $V_{\pi}(\bel) = \int_{\st\in\stSpace} \bel(\st)\alpha_{\sigma}(\st)\de\st$. Note that \eref{eq:alpha_s} is defined with respect to POMDP \pomdp. Analogously we define the linearized $\alpha$-function $\widehat{\alpha}_{\sigma}$ with respect to the linearized POMDP \linPomdp by replacing the transition and observation functions in \eref{eq:alpha_s} with their linearized versions.

Now, suppose that for a given belief \bel,  $\sigma^* = \argsup_{\sigma\in\Gamma} \int_{\st\in\stSpace}\bel(\st)\alpha_{\sigma}(\st)\de\st$ and $\widehat{\sigma}^* = \argsup_{\sigma\in\Gamma}\int_{\st\in\stSpace}\bel(\st)\widehat{\alpha}_{\sigma}(\st)\de\st$. $\sigma^*$ and $\widehat{\sigma}^*$ represent the policies $\pi^*$ and $\widehat{\pi}^*$ that are optimal at \bel for POMDP \pomdp and \linPomdp respectively. For any $\st\in\stSpace$ we have that $\alphaFunctPolComp{\widehat{\sigma}^*}{\st} \geq \linAlphaFunctPolComp{\widehat{\sigma}^*}{\st} - \left |\alphaFunctPolComp{\widehat{\sigma}^*}{\st} - \linAlphaFunctPolComp{\widehat{\sigma}^*}{\st} \right |$ and $\linAlphaFunctPolComp{\sigma^*}{\st} \geq \alphaFunctPolComp{\sigma^*}{\st} - \left | \alphaFunctPolComp{\sigma^*}{\st} - \linAlphaFunctPolComp{\sigma^*}{\st}\right |$. Therefore 
\begin{align}\label{eq:geq_1}
\intS \belS{\st} \alphaFunctPolComp{\widehat{\sigma}^*}{\st} \de\st \geq &\intS \belS{\st} \linAlphaFunctPolComp{\widehat{\sigma}^*}{\st} \de\st\nonumber \\ & - \intS \belS{\st} \left | \alphaFunctPolComp{\widehat{\sigma}^*}{\st} - \linAlphaFunctPolComp{\widehat{\sigma}^*}{\st} \right | \de\st
\end{align}

and 
\begin{align}\label{eq:geq_2}
\intS \belS{\st} \linAlphaFunctPolComp{\sigma^*}{\st} \de\st \geq &\intS \belS{\st}\alphaFunctPolComp{\sigma^*}{\st} \de\st\nonumber \\&- \intS \belS{\st}\left |\alphaFunctPolComp{\sigma^*}{\st} - \linAlphaFunctPolComp{\sigma^*}{\st} \right |\de\st
\end{align}

Since $\widehat{\sigma}^*$ is the optimal conditional plan for POMDP \linPomdp at \bel, we also know that
\begin{equation}\label{eq:geq_3}
\intS \belS{\st} \linAlphaFunctPolComp{\widehat{\sigma}^*}{\st} \de\st \geq \intS \belS{\st} \linAlphaFunctPolComp{\sigma^*}{\st} \de\st
\end{equation}

From \eref{eq:geq_1}, \eref{eq:geq_2} and \eref{eq:geq_3} it immediately follows that
\begin{alignat}{2}\label{eq:geq_4}
\intS \belS{\st} \alphaFunctPolComp{\widehat{\sigma}^*}{\st} \de\st \geq & &&\intS \belS{\st} \alphaFunctPolComp{\sigma^*}{\st} \de\st \nonumber \\ & && - 2 \intS \belS{\st}\sup_{\sigma\in\Gamma}\left |\alphaFunctPolComp{\sigma}{\st} - \linAlphaFunctPolComp{\sigma}{\st} \right |\de\st \nonumber \\
V_{\widehat{\pi}^*}(b) \geq& && V_{\pi^*}(b) \nonumber \\ & &&- 2 \intS \belS{\st}\sup_{\sigma\in\Gamma}\left |\alphaFunctPolComp{\sigma}{\st} - \linAlphaFunctPolComp{\sigma}{\st} \right | \de\st
\end{alignat}

Before we continue, we first have to show the following Lemma:
\begin{lemma}\label{lem:lem0}
Let $R_m = \max\{\left |R_{min} \right |, R_{max}\}$, where $R_{min} = \min_{s,a} R(s, a)$ and $R_{max} = \max_{s,a} R(s, a)$. For any conditional plan $\sigma\in\Gamma$ and any $\st \in \stSpace$, the absolute difference between the original and linearized $\alpha$-functions is upper bounded by
\begin{align}
\left | \alphaFunctPolComp{\sigma}{\st} - \linAlphaFunctPolComp{\sigma}{\st} \right | \leq 2\gamma\frac{R_{m}}{(1-\gamma)^2} \nm{\pomdp}{\linPomdp}\nonumber
\end{align}
\end{lemma}

The proof of \lref{lem:lem0} is presented in the Appendix~\ref{ssec:lemma_0_proof}.
 
Using the result of \lref{lem:lem0}, we can now conclude the proof for \thref{th:valUpperBound}. Substituting the upper bound derived in \lref{lem:lem0} into the right-hand side of \eref{eq:geq_4} and re-arranging the terms gives us
\begin{equation}
V_{\pi^*}(\bel) - V_{\widehat{\pi}^*}(\bel) \leq 4\gamma\frac{R_{m}}{(1-\gamma)^2} \nm{\pomdp}{\linPomdp}
\end{equation}

which is what we are looking for. $\square$

\section{Approximating \monNameAbbr}\label{ssec:monApprox}
Now, the question is how can we compute \monNameAbbr sufficiently fast, so that this measure can be used as a heuristic during on-line planning to decide when a linearization-based solver will likely yield a good policy and when a general solver should be used. Unfortunately, such a computation is often infeasible when the planning time per step is limited. Therefore, we  approximate \monNameAbbr off-line and re-use the results during run-time. Here we discuss how to approximate the transition component \nmSymT of \monNameAbbr, however, the same method applies to the observation component \nmSymZ.

Let us first rewrite the transition component of \nmSymT as
\begin{align}\label{e:snmTransCompRe}
\nmSymT &= \sup_{\st \in \stSpace} \nmSymT(s)\nonumber \\&= \sup_{\st\in\stSpace}\sup_{\act\in\actSpace}D_{TV}(\transComp, \linTransComp)
\end{align}
where $\nmSymT(s)$ is the transition component of \monNameAbbr, given a particular state. To approximate \nmSymT, we replace \stSpace in \eref{e:snmTransCompRe} by a sampled representation of \stSpace, which we denote as $\tilde{\stSpace}$. The value $\nmSymT(s)$ is then evaluated for each $\st\in\tilde{\stSpace}$ off-line, and the results are saved in a lookup-table. This lookup-table can then be used during run-time to get a local approximation of \nmSymT around the current belief.

The first question that arises is, how do we efficiently sample the state space? A naive approach would be to employ a simple uniform sampling strategy. However, for large state spaces this is often wasteful, because for motion planning problems, large portions of the state space are often irrelevant since they either can't be reached from the initial belief or are unlikely to be traversed by the robot during run-time. A better strategy is to consider only the subset of the state space that is reachable from the support set of the initial belief under any policy, denoted as $\stSpace_{\belInit}$. To sample from $\stSpace_{\belInit}$, we use a simple but effective method: Assuming deterministic dynamics, we solve the motion planning problem off-line using kinodynamic RRTs and use the nodes in the RRT-trees as a sampled representation of $\stSpace_{\belInit}$. In principle any deterministic sampling-based motion planner can be used to generate samples from $\stSpace_{\belInit}$, however, in our case RRT is a particularly suitable due to its space-filling property \ccite{kuffner2011space}. Note that RRT generates states according to a deterministic transition function only. If required, one could also generate additional samples according to the actual stochastic transition function of the robot. However, in our experiments the state samples generated by RRT were sufficient.  

The second difficulty in approximating $\nmSymT(s)$ is the computation of the supremum over the action space. Similar to restricting the approximation to a discrete set of states reachable from the initial belief, we can impose a discretization on the action space which leaves us with a maximization over discrete actions, denoted as $\tilde{\actSpace}$. Using the set $\tilde{A}$, we  approximate \eref{e:snmTransCompRe} for each state in $\tilde{\stSpace}_{\belInit}$ ---the sampled set of $\stSpace_{\belInit}$--- as follows:
Given a particular state $\st \in \tilde{\stSpace}_{\belInit}$ and action $\act \in \tilde{\actSpace}$, we draw $n$ samples from the original and linearized transition function and construct a multidimensional histogram from both sample sets. In other words, we discretize the distributions that follow from the original and linearized transition function, given a particular state and action. Suppose the histogram consists of $k$ bins. The value $\nmSymT(\st, \act)$ is then approximated as
\begin{equation}\label{eq:smmTransCompStAct}
  \nmSymT(\st, \act) \approx \frac{1}{2} \sum_{i = 1}^{k} \left |p_i - \widehat{p}_i \right |
\end{equation}
where $p_i = \frac{n_i}{\sum_{j=1}^k n_j}$ and $n_i$ is the number of states inside bin $i$ sampled from the original transition function, while $\widehat{p}_i = \frac{\widehat{n}_i}{\sum_{j=1}^k \widehat{n}_j}$ and $\widehat{n}_i$ is the number of states inside bin $i$ sampled from the linearized transition function. The right-hand side of \eref{eq:smmTransCompStAct} is simply the definition of the total variation distance between two discrete distributions.

By repeating the above process for each action in $\tilde{A}$ and taking the maximum, we end up with an approximation of $\nmSymT(s)$. This procedure is repeated for every state in the set $\tilde{\stSpace}_{\belInit}$. As a result we get a lookup-table, assigning each state in $\tilde{\stSpace}_{\belInit}$ an approximated value of $\nmSymT(s)$.

During planning, we can use the lookup-table and a sampled representation of a belief \bel to approximate \monNameAbbr at \bel. Suppose $\tilde{\bel}$ is the sampled representation of \bel (e.g., a particle set), then for each state $s \in \tilde{\bel}$,  we take the state $\st_{near} \in \tilde{\stSpace}_{\belInit}$ that is nearest to $s$, and assign $\nmSymT(s) = \nmSymT(\st_{near})$. The maximum SNM value $\max_{s \in \tilde{\bel}} \nmSymT(s)$ gives us an approximation of the transition component of \monNameAbbr with respect to the belief \bel.

Clearly this approximation method assumes that states that are close together should yield similar values for \monNameAbbr. At first glance this is a very strong assumption. In the vicinity of obstacles or constraints, states that are close together could potentially yield very different \monNameAbbr values. However, we will now show that under mild assumptions, pairs of states that are elements within certain subsets of the state space indeed yield similar \monNameAbbr values.

Consider a partitioning of the state space into a finite number of local-Lipschitz subsets $S_i$ that are defined as follows:
\begin{definition}\label{d:partition}
Let \stSpace be a metric space with distance metric $D_{\stSpace}$. $\stSpace_i$ is called a local-Lipschitz subset of $\stSpace$ if for any $\st_1, \st_2 \in \stSpace_i$, any $\stp \in \stSpace$ and any $\act \in \actSpace: \left |\trans(\st_1, \act, \stp) - \trans(\st_2, \act, \stp) \right | \leq C_{\trans_i}D_{\stSpace}(\st_1, \st_2)$ and $\left |\linTrans(\st_1, \act, \stp) - \linTrans(\st_2, \act, \stp) \right | \leq C_{\linTrans_i}D_{\stSpace}(\st_1, \st_2)$, where $C_{\trans_i} \geq 0$ and $C_{\linTrans_i} \geq 0$ are finite local-Lipschitz constants
\end{definition}
In other words, $\stSpace_i$ are subsets of $\stSpace$ in which the original and linearized transition functions are Lipschitz continuous with Lipschitz constants $C_{\trans_i}$ and $C_{\linTrans_i}$. With this definition at hand, we can now show the following lemma:
\begin{lemma}\label{lem:lemApprox}
Let \stSpace be a $n-dimensional$ metric space with distance metric $D_{\stSpace}$ and assume \stSpace is normalized to $\left [0, 1 \right ]^n$. Furthermore let $\stSpace_i$ be a local-Lipschitz subset of \stSpace, then
\begin{equation}
\left | \nmSymT(\st_1) - \nmSymT(\st_2) \right | \leq \frac{1}{2} \sqrt{n} D_{\stSpace}(\st_1, \st_2) \left [C_{\trans_i} + C_{\linTrans_i} \right ] \nonumber
\end{equation}
 for any $\st_1, \st_2 \in S_i$
\label{l:subsetLipschitz}
\end{lemma}
The proof for this Lemma is presented in \appref{ssec:proofLem4}. This Lemma indicates that the difference between the \monNameAbbr values for two states from the same local-Lipschitz subset $\stSpace_i$ depends only on the distance $D_{\stSpace}$ between them, since $C_{\trans_i}$ and $C_{\linTrans_i}$ are constant for each subset $\stSpace_i$. Thus, as the distance between two states converges towards zero, the \monNameAbbr value difference converges towards zero as well. This implies that we can approximate SNM for a sparse, sampled representation of $\stSpace_{\belInit}$ and re-use these approximations on-line with a small error, without requiring an explicit representation of the $\stSpace_i$ subsets.

\section{SNM-Planner: An Application of \monNameAbbr for Planning}
\label{sec:method}
\nop is an on-line planner that uses \monNameAbbr as a heuristic to decide whether a general, or a linearization-based POMDP solver should be used to compute the policy from the current belief. The general solver used is Adaptive Belief Tree (ABT)\ccite{Kur13:An}, while the linearization-based method called Modified High Frequency Replanning (MHFR), which is an adaptation of HFR\ccite{Sun15:High}. HFR is designed for chance-constraint POMDPs, i.e., it explicitly minimizes the collision probability, while MHFR is a POMDP solver where the objective is to maximize the expected total reward. An overview of \nop is shown in \aref{alg:smnd}. During run-time, at each planning step, \nop computes a local approximation of \monNameAbbr around the current belief $\bel_i$ (line 5). If this value is smaller than a given threshold, \nop uses MHFR to compute a policy from the current belief, whereas ABT is used when the value exceeds the threshold (line 8-12). The robot then executes an action according the computed policy (line 13) and receives and observation (line 14). Based on the executed action and perceived observation, we update the belief (line 15). \nop represents beliefs as sets of particles and updates the belief using a SIR particle filter\ccite{arulampalam2002tutorial}. Note that MHFR assumes that beliefs are multivariate Gaussian distributions. Therefore, in case MHFR is used for the policy computation, we compute the first two moments (mean and covariance) of the particle set to obtain a multivariate Gaussian approximation of the current belief. The process then repeats from the updated belief until the robot has entered a terminal state (we assume that we know when the robot enters a terminal state) or until a maximum number of steps is reached. 

In the following two subsections we provide a brief an overview of the two component planners ABT and MHFR.

\begin{algorithm}
\caption{\nop (initial belief \belInit, \monNameAbbr threshold $\mu$, max. planning time per step $t$, max. number of steps $N$)}\label{alg:smnd}
\begin{algorithmic}[1]
\State InitializeABT(\pomdp)
\State InitializeMHFR(\pomdp)
\State $i=0$, $\bel_i = \belInit$, terminal = False
\While{terminal is False and $i < N$}
\State $\widehat{\Psi} =\ $approximateSNM($\bel_i$)
\State $t_p = t - t_a$
\Comment $t_{a}$ is the time the algorithm takes to approximate \monNameAbbr
\If{$\widehat{\Psi} < \mu$}	
	\State $a =\ $MHFR($\bel_i$, $t_p$)  
\Else
   \State $a =\ $ABT($\bel_i$, $t_p$)
\EndIf
\State terminal = executeAction($a$)
\State $o =\ $get observation
\State $b_{i+1} = \tau(\bel_i, a, o)$
\State $i = i + 1$
\EndWhile
\end{algorithmic}
\end{algorithm}

\subsection{Adaptive Belief Tree (ABT)}\label{ssec:ABT}
ABT is a general and anytime on-line POMDP solver based on Monte-Carlo-Tree-Search (MCTS). ABT updates (rather than recomputes) its policy at each planning step. To update the policy for the current belief, ABT iteratively constructs and maintains a belief tree, a tree whose nodes are beliefs and whose edges are pairs of actions and observations. ABT evaluates sequences of actions by sampling episodes, that is, sequences of state-–action-–observation-–reward tuples, starting from the current belief. Details of ABT can be found in\ccite{Kur13:An}.

\subsection{Modified High-Frequency Replanning (MHFR)}\label{ssec:MHFR}
The main difference between HFR and MHFR is that HFR is designed for chance constraint POMDP, i.e., it explicitly minimizes the collision probability, while MHFR is a POMDP solver, whose objective is to maximize the expected total reward. Similar to HFR, MHFR approximates the current belief by a multivariate Gaussian distribution. To compute the policy from the current belief, MHFR samples a set of trajectories from the mean of the current belief to a goal state using multiple instances of RRTs\ccite{kuffner2011space} in parallel. It then computes the expected total discounted reward of each trajectory by tracking the beliefs around the trajectory using a Kalman Filter, assuming maximum-likelihood observations. The policy then becomes the first action of the trajectory with the highest expected total discounted reward. After executing the action and perceiving an observation, MHFR updates the belief using an Extended Kalman Filter. The process then repeats from the updated belief. To increase efficiency, MHFR additionally adjusts the previous trajectory with the highest expected total discounted reward to start from the mean of the updated belief and adds this trajectory to the set of sampled trajectories. More details on HFR and precise derivations of the method are available in\ccite{Sun15:High}.

\section{Experiments and Results}
The purpose of our experiments is two-fold: To test the applicability of \monNameAbbr to motion planning under uncertainty problems and to test \nop. For our first objective, we compare \monNameAbbr with a modified version of the Measure of Non-Gaussianity (\monGAbbr)\ccite{Dun13:Nonlinearity}. Details on this measure are in \sref{ssec:mong}. We evaluate both measures using two robotic systems, a car-like robot with 2$^{nd}$-order dynamics and a torque-controlled 4DOFs manipulator, where both robots are subject to increasing uncertainties and increasing numbers of obstacles in the operating environment. Furthermore we test both measures when the robots are subject to highly non-linear collision dynamics and different observation models. Details on the robot models are presented in \sref{ssec:robot_models}, whereas the evaluation experiments are presented in \sref{ssec:testing_snm}.

To test \nop we compare it with ABT and MHFR on three problem scenarios, including a torque-controlled 7DOFs manipulator operating inside a 3D office environment. Additionally we test how sensitive \nop is to the choice of the \monNameAbbr-threshold. The results for these experiments are presented in \sref{ssec:testing_snm-planner}.

All problem environments are modelled within the OPPT framework\ccite{hoerger2018software}. The solvers are implemented in C++. For the parallel construction of the RRTs in MHFR, we utilize 8 CPU cores throughout the experiments. All parameters are set based on preliminary runs over the possible parameter space, the parameters that generate the best results are then chosen to generate the experimental results.

\subsection{\monG}\label{ssec:mong}
The Measure of Non-Gaussianity (\monGAbbr) proposed in\ccite{Dun13:Nonlinearity} is based on the negentropy between the PDF of a random variable and its Gaussian approximation. Consider an  $n$-dimensional random variable $X$ distributed according to PDF $p(x)$. Furthermore, let \lin{X} be a Gaussian approximation of $X$ with PDF $\widehat{p}(x)$, such that $\lin{X} \sim N(\mu , \Sigma_x)$, where $\mu$ and $\Sigma_x$ are the first two moments of $p(x)$. The negentropy between $p$ and \lin{p} (denoted as \J{p}{\lin{p}}) is then defined as
\begin{equation}
\J{p}{\lin{p}} = H(\lin{p}) - H(p)
\end{equation}
where
\begin{equation}\label{e:entropies}
\begin{split}
\entr{\lin{p}} &= \frac{1}{2} ln \left [(2 \pi e)^n \left |det(\Sigma_x) \right | \right ] \\
\entr{p} &= - \int p(x) \ln p(x) dx
\end{split}
\end{equation}
are the differential entropies of $p$ and $\lin{p}$ respectively.
A (multivariate) Gaussian distribution has the largest differential entropy amongst all distributions with equal first two moments, therefore \J{p}{\lin{p}} is always non-negative. In practice, since the PDF $p(x)$ is not known exactly in all but the simplest cases, \entr{p} has to be approximated. 

In\ccite{Dun13:Nonlinearity} this measure has originally been used to assess the non-linearity of passive systems. Therefore, in order to achieve comparability with \monNameAbbr, we need to extend the Non-Gaussian measure to general active stochastic systems of the form $s_{t+1} = f(s_t, a_t, v_t)$. We do this by evaluating the non-Gaussianity of distribution that follow from the transition function \transComp given state $s$ and action $a$. In particular for a given $s$ and $a$, we can find a Gaussian approximation of \transComp (denoted by \transCompGauss) by calculating the first two moments of the distribution that follows from \transComp. 

Using this Gaussian approximation, we define the \monG as
\begin{align}
&\monGAbbr(\trans, \transGauss) = \nonumber\\ &\sup_{\st \in \stSpace, \act \in \actSpace} \left [H(\transCompGauss) - H(\transComp)\right ]
\end{align}

Similarly we can compute the \monG for the observation function:
\begin{align}
&\monGAbbr(\obsF, \obsFGauss) = \nonumber\\ &\sup_{\st \in \stSpace, \act \in \actSpace} \left [H(\obsFCompGauss) - H(\obsFComp) \right ]
\end{align}

where \obsFGauss is a Gaussian approximation of \obsF.

In order to approximate the entropies $\entr{\transComp)}$ and \entr{\obsFComp}, we are using a similar histogram-based approach as discussed in \sref{ssec:monApprox}. The entropy terms for the Gaussian approximations can be computed in closed form, according to the first equation in \eref{e:entropies}\ccite{Ahmed89Entropy}.

\subsection{Robot Models}\label{ssec:robot_models}
\subsubsection{4DOFs-Manipulator. }\label{sssec:4DOFManipulator}
The 4DOFs-manipulator consists of 4 links connected by 4 torque-controlled revolute joints. The first joint is connected to a static base. In all problem scenarios the manipulator must move from a known initial state to a state where the end-effector lies inside a goal region located in the workspace of the robot, while avoiding collisions with obstacles the environment is populated with.

The state of the manipulator is defined as $\st=(\theta, \dot{\theta}) \in \mathbb{R}^{8}$, where $\theta$ is the vector of joint angles, and $\dot{\theta}$ the vector of joint velocities. Both joint angles and joint velocities are subject to linear constraints: The joint angles are constrained by $(-3.14, 3.14)rad$, whereas the joint velocities are constrained by $(6,\allowbreak 2,\allowbreak 2,\allowbreak 2)rad/s$ in each direction. Each link of the robot has a mass of $1kg$.

The control inputs of the manipulator are the joint torques, where the maximum joint torques are $(20,\allowbreak 20,\allowbreak 10,\allowbreak 5)Nm/s$ in each direction. Since ABT assumes a discrete action space, we discretize the joint torques for each joint using the maximum torque in each direction, which leads to 16 actions.

The dynamics of the manipulator is defined using the well-known Newton-Euler formalism\ccite{spong06:RobotModelling}. For both manipulators we assume that the input torque for each joint is affected by zero-mean additive Gaussian noise. Note however, even though the error is Gaussian, due to the non-linearities of the motion dynamics the beliefs will not be Gaussian in general. Since the transition dynamics for this robot are quite complex, we assume that the joint torques are applied for 0.1s and we use the ODE physics engine\ccite{drumwright2010:extending} for the numerical integration of the dynamics, where the discretization (\ie\ $\delta t$) of the integrator is set to $\delta t = 0.004s$.

The robot is equipped with two sensors: The first sensor measures the position of the end-effector inside the robot's workspace, whereas the second sensor measures the joint velocities. Consider a function $g: \mathbb{R}^{8} \mapsto \mathbb{R}^3$ that maps the state of the robot to an end-effector position inside the workspace, then the observation model is defined as 
\begin{equation}\label{eq:obs4DOF}
o = [g(s), \dot{\theta}]^T + w
\end{equation}
where $w_t$ is an error term drawn from a zero-mean multivariate Gaussian distribution with covariance matrix $\Sigma_w$.

The initial state of the robot is a state where the joint angles and velocities are zero.

When the robot performs an action where it collides with an obstacle it enters a terminal state and receives a penalty of -500. When it reaches the goal area it also enters a terminal state, but receives a reward of 1,000. To encourage the robot to reach the goal area quickly, it receives a small penalty of -1 for every other action. 
\subsubsection{7DOFs Kuka iiwa manipulator. }\label{sssec:7DOFManipulator}
The 7DOFs Kuka iiwa manipulator is very similar to the 4DOFs-manipulator. However, the robot consists of 7 links connected via 7 revolute joints. We set the POMDP model to be similar to that of the 4DOFs-manipulator, but expand it to handle 7DOFs. For this robot, the joint velocities are constrained by $(3.92,\allowbreak 2.91,\allowbreak 2.53,\allowbreak 2.23,\allowbreak 2.23,\allowbreak 2.23,\allowbreak 1.0)rad/s$ in each direction and the link masses are $(4,\allowbreak 4,\allowbreak 3,\allowbreak 2.7,\allowbreak 1.7,\allowbreak 1.8,\allowbreak 0.3)kg$. Additionally, the torque limits of the joints are $(25,\allowbreak 20,\allowbreak 10,\allowbreak 10,\allowbreak 5,\allowbreak 5,\allowbreak 0.5)Nm/s$ in each direction. For ABT we use the same discretization of the joint torques as in the 4DOFs-manipulator case, \ie we use the maximum torque per joint in each direction, resulting in 128 actions. Similarly to the 4DOFs-manipulator, we assume that the input torques are applied for 0.1s and we use the ODE physics engine with an integration step size of 0.004s to simulate the transition dynamics. The observation and reward models are the same as for the 4DOFs-manipulator. The initial joint velocities are all zero and almost all joint angles are zero too, except for the second joint, for which the initial joint angle is $-1.5rad$. \fref{f:scenarioCompStudy}(c) shows the Kuka manipulator operating inside an office scenario.

\subsubsection{Car-like robot. }\label{sssec:SecOrderCar. }
A nonholonomic car-like robot of size ($0.12\times0.07\times0.01$) drives on a flat xy-plane inside a 3D environment populated by obstacles The robot must drive from a known start state to a position inside a goal region without colliding with any of the obstacles. The state of the robot at time t is defined as a 4D vector $s_t = (x_t, y_t, \theta_t, \upsilon_t) \in \mathbb{R}^4$, where $x_t, y_t \in [-1, 1]$ is the position of the center of the robot on the $xy$-plane, $\theta_t \in [-3.14, 3.14]rad$ the orientation and $\upsilon_t \in [-0.2, 0.2]$ is the linear velocity of the robot. The initial state of the robot is $(−0.7,−0.7,1.57rad,0)$ while the goal region is centered at $(0.7,0.7)$ with radius $0.1$. The control input at time $t$, $a_t = (\alpha_t, \phi_t)$ is a 2D real vector consisting of the acceleration $\alpha \in [-1, 1]$ and the steering wheel angle $\phi_t \in [-1rad, 1rad]$. The robot’s dynamics is subject to control noise $v_t = (\tilde{\alpha}_t, \tilde{\phi}_t) \sim N(0, \Sigma_v)$. The robot’s transition model is
\begin{equation}\label{eq:carDynamics}
  s_{t+1} = f(s_t, a_t, v_t) = \begin{bmatrix}
x_t + \Delta t \upsilon_t \cos \theta_t\\ 
y_t + \Delta t \upsilon_t \sin \theta_t\\ 
\theta_t + \Delta t \tan(\phi_t + \tilde{\phi}_t) / 0.11\\ 
\upsilon_t + \Delta t (\alpha_t + \tilde{\alpha}_t)
\end{bmatrix}
\end{equation}
where $\Delta t = 0.3s$ is the duration of a timestep and the value $0.11$ is the distance between the front and rear axles of the wheels. 

This robot is equipped with two types of sensors, a localization sensor that receives a signal from two beacons that are located at $(\hat{x}_1, \hat{y}_1)$ and $(\hat{x}_2, \hat{y}_2)$. The second sensor is a velocity sensor mounted on the car. With these two sensors the observation model is defined as
\begin{equation}\label{e:obstFunctCarMazeAdditive}
o_t = \begin{bmatrix}
\frac{1}{((x_t - \hat{x}_1)^2 + (y_t - \hat{y}_1)^2 + 1)}\\ 
\frac{1}{((x_t - \hat{x}_2)^2 + (y_t - \hat{y}_2)^2 + 1)}\\ 
v_t
\end{bmatrix} + w_t
\end{equation}

where $w_t$ is an error vector drawn from a zero-mean multivariate Gaussian distribution with covariance matrix $\Sigma_w$.

Similar to the manipulators described above, the robot receives a penalty of -500 when it collides with an obstacle, a reward of 1,000 when reaching the goal area and a small penalty of -1 for any other action.

\subsection{Testing \monNameAbbr}\label{ssec:testing_snm}
In this set of experiments we want to understand the performance of \monNameAbbr compared to \monGAbbr in various scenarios. In particular, we are interested in the effect of increasing uncertainties and the effect that obstacles have on the effectiveness of \monNameAbbr, and if these results are consistent with the performance of a general solver relative to a linearization-based solver. Additionally, we want to see how highly-nonlinear collision dynamics and different observation models -- one with additive Gaussian noise and non-additive Gaussian noise -- affect our measure. 
For the experiments with increasing motion and sensing errors, recall from \sref{ssec:robot_models} that the control errors are drawn from zero-mean multivariate Gaussian distributions with covariance matrices $\Sigma_v$. We define the control errors (denoted as \pet) to be the standard deviation of these Gaussian distributions, such that $\Sigma_v = \pet^2 \times \mathds{1}$. Similarly for the covariance matrices of the zero-mean multivariate Gaussian sensing errors, we define the observation error as \pez, such that $\Sigma_w = \pez^2 \times \mathds{1}$. Note that during all the experiments, we use normalized spaces, which means that the error vectors affect the normalized action and observation vectors.
For \monNameAbbr and \monGAbbr we first generated 100,000 state samples for each scenario, and computed a lookup table for each error value off-line, as discussed in \sref{ssec:monApprox}. Then, during run-time we calculated the average approximated SNM and MonG values. 
\subsubsection{Effects of increasing uncertainties in cluttered environments. }\label{ssec:increasingly_uncertain}
\begin{figure*}
\centering
\begin{tabular}{c@{\hspace*{5pt}}c@{\hspace*{5pt}}c@{\hspace*{5pt}}}
\includegraphics[height=4cm]{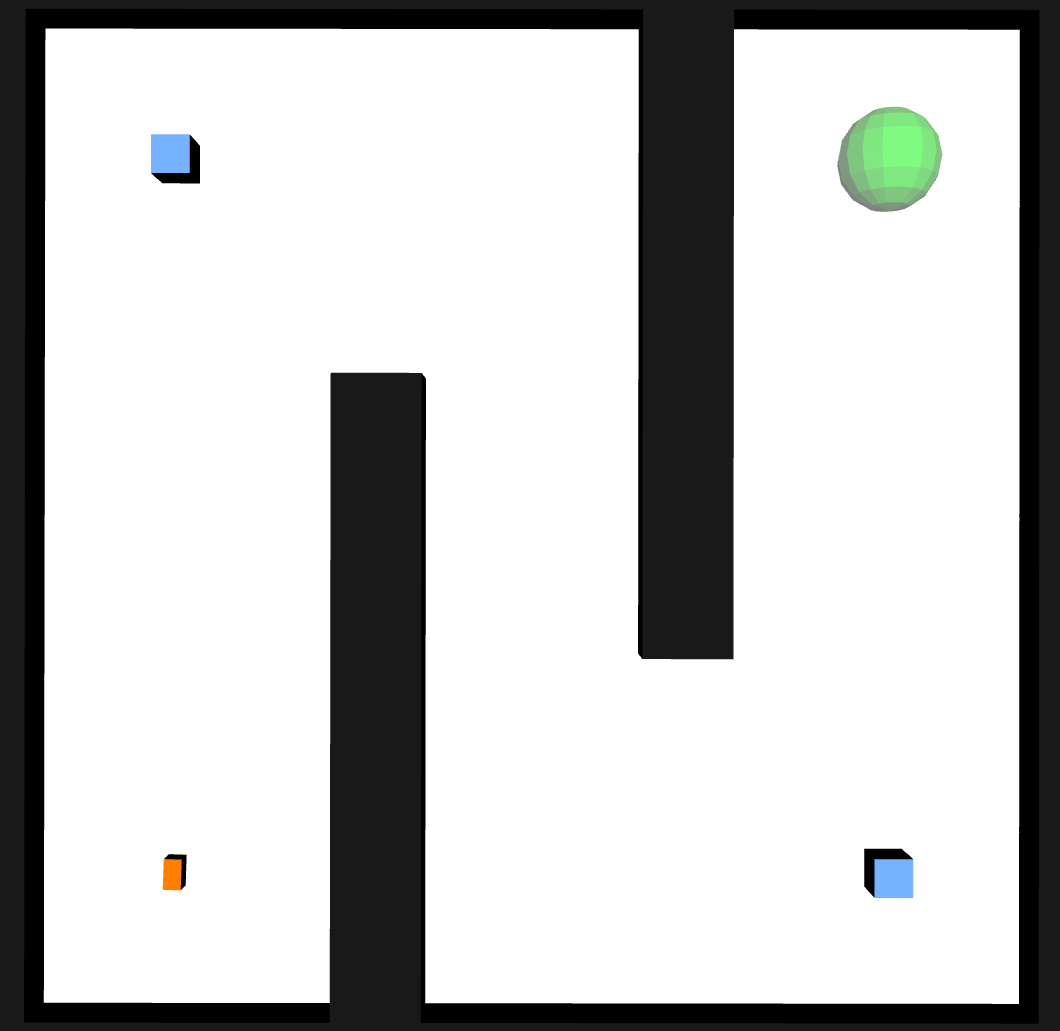} &
\includegraphics[height=4cm]{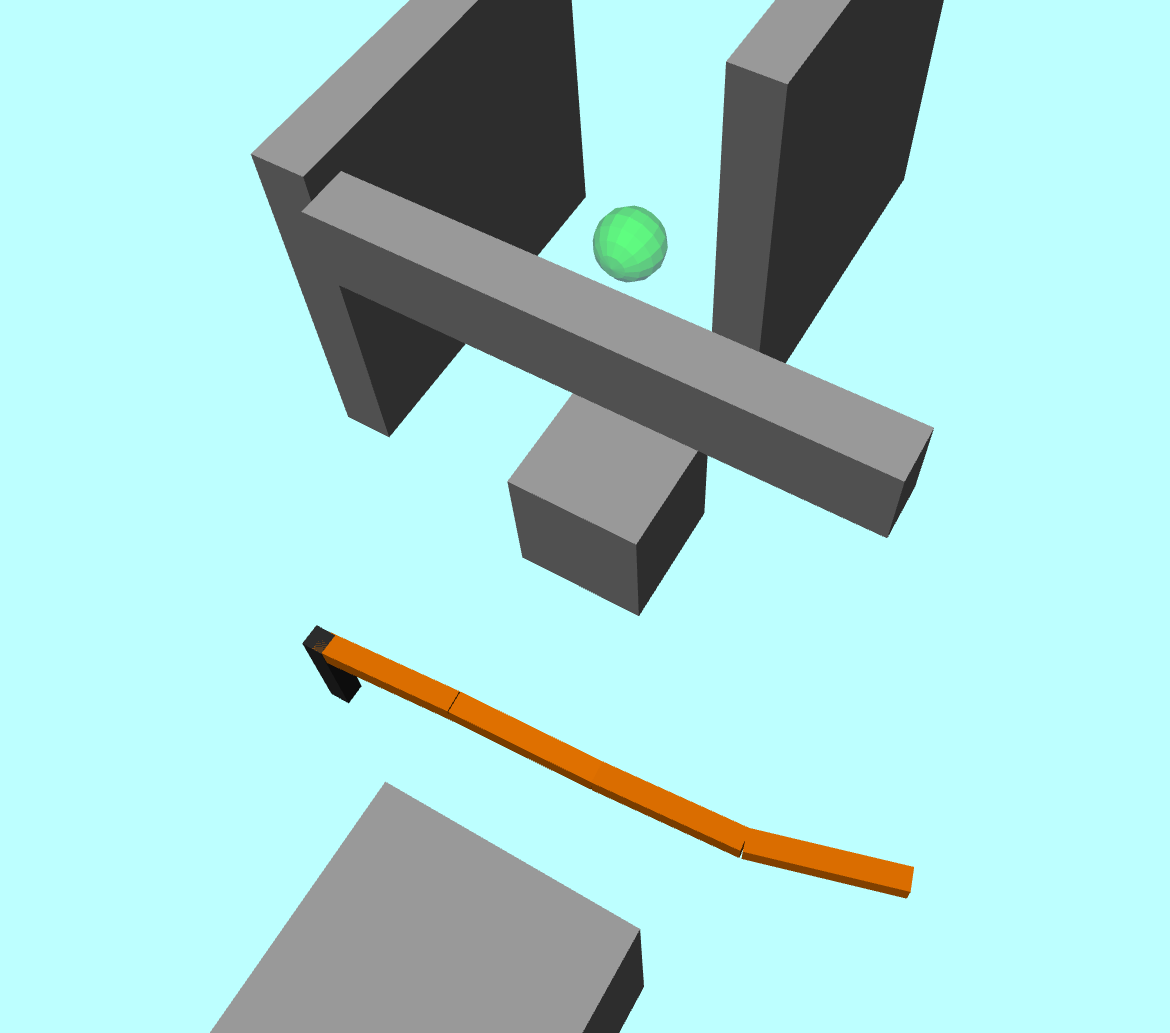} &
\includegraphics[height=4cm]{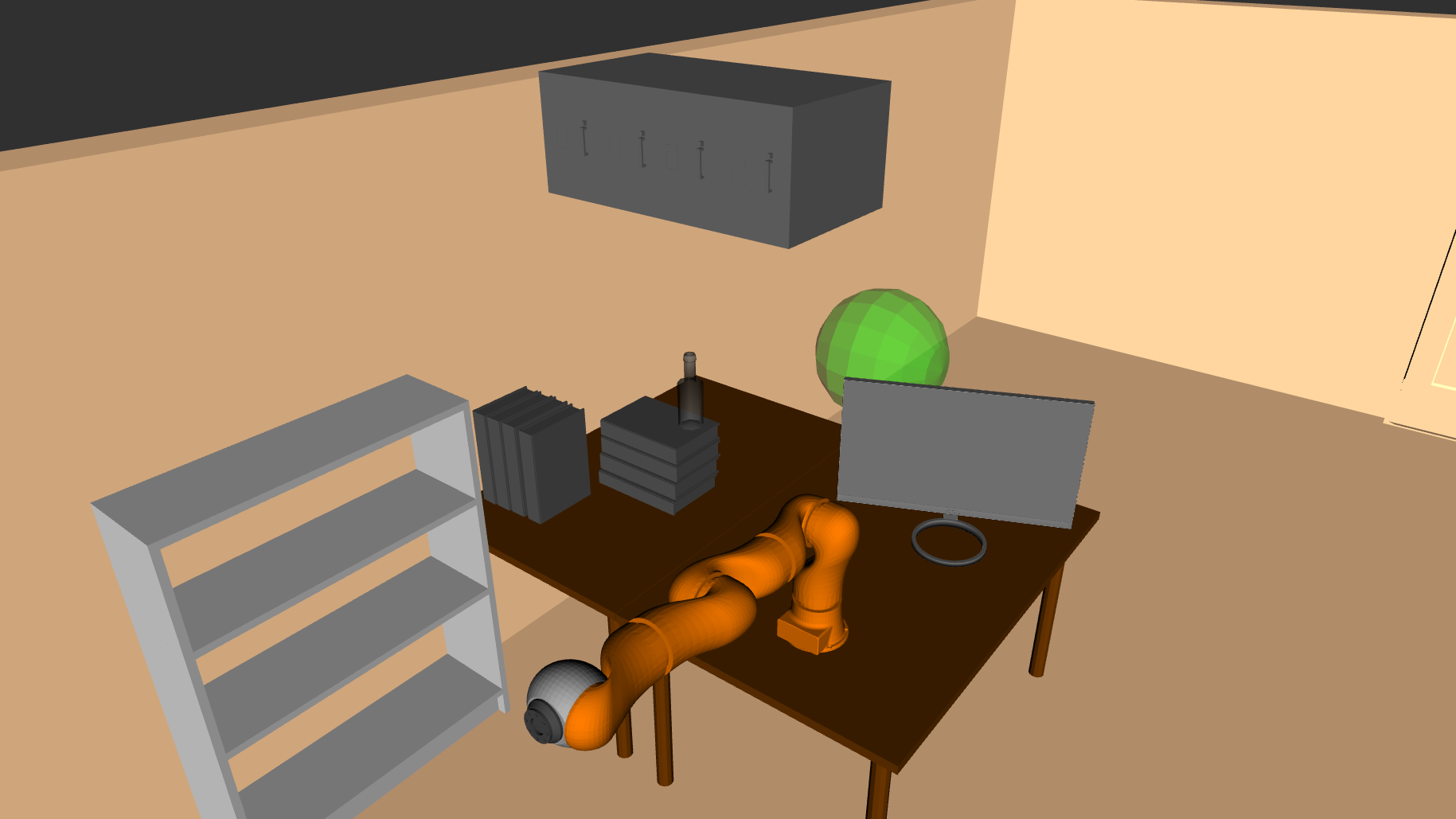} \\
(a) Maze & (b) Factory & (c) KukaOffice
\end{tabular}
\caption{Test scenarios for the different robots. The objects colored black and gray are obstacles, while the green sphere is the goal region. (a) The Maze scenario for the car-like robot. The blue squares represents the beacons, while the orange square at the bottom left represents the initial state. (b) The 4DOFs-manipulator scenario. (c) The KukaOffice scenario}
\label{f:scenarioCompStudy}
\end{figure*}

To investigate the effect of increasing control and observation errors to \monNameAbbr, \monGAbbr and the two solvers ABT and MHFR in cluttered environments, we ran a set of experiments where the 4DOFs-manipulator and the car-like robot operate in empty environments and environments with obstacles,  with increasing values of \pet and \pez, ranging between $0.001$ and $0.075$. The environments with obstacles are the Factory and Maze environments shown in \fref{f:scenarioCompStudy}(a) and (b). For each scenario and each control-sensing error value (we set $\pet = \pez$), we ran 100 simulation runs using ABT and MHFR respectively with a planning time of 2s per step. 

The average values for \monNameAbbr and \monGAbbr and the relative value differences between ABT and MHFR in the empty environments are presented in \tref{t:measureCompareEmpty}.
The results show that for both scenarios \monNameAbbr and \monGAbbr are sensitive to increasing transition and observation errors. This resonates well with the relative value difference between ABT and MHFR. The more interesting question is now, how sensitive are both measures to obstacles in the environment? \tref{t:measureCompareClutter}(a) and (b) shows the results for the Factory and the Maze scenario respectively. It is evident that \monNameAbbr increases significantly compared to the empty environments, whereas \monGAbbr is almost unaffected. Overall obstacles increase the relative value difference between ABT and MHFR, except for large uncertainties in the Maze scenario. This indicates that MHFR suffers more from the additional non-linearities that obstacles introduce. \monNameAbbr is able to capture these effects well.

\begin{figure}
\centering
\includegraphics[width=0.485\textwidth]{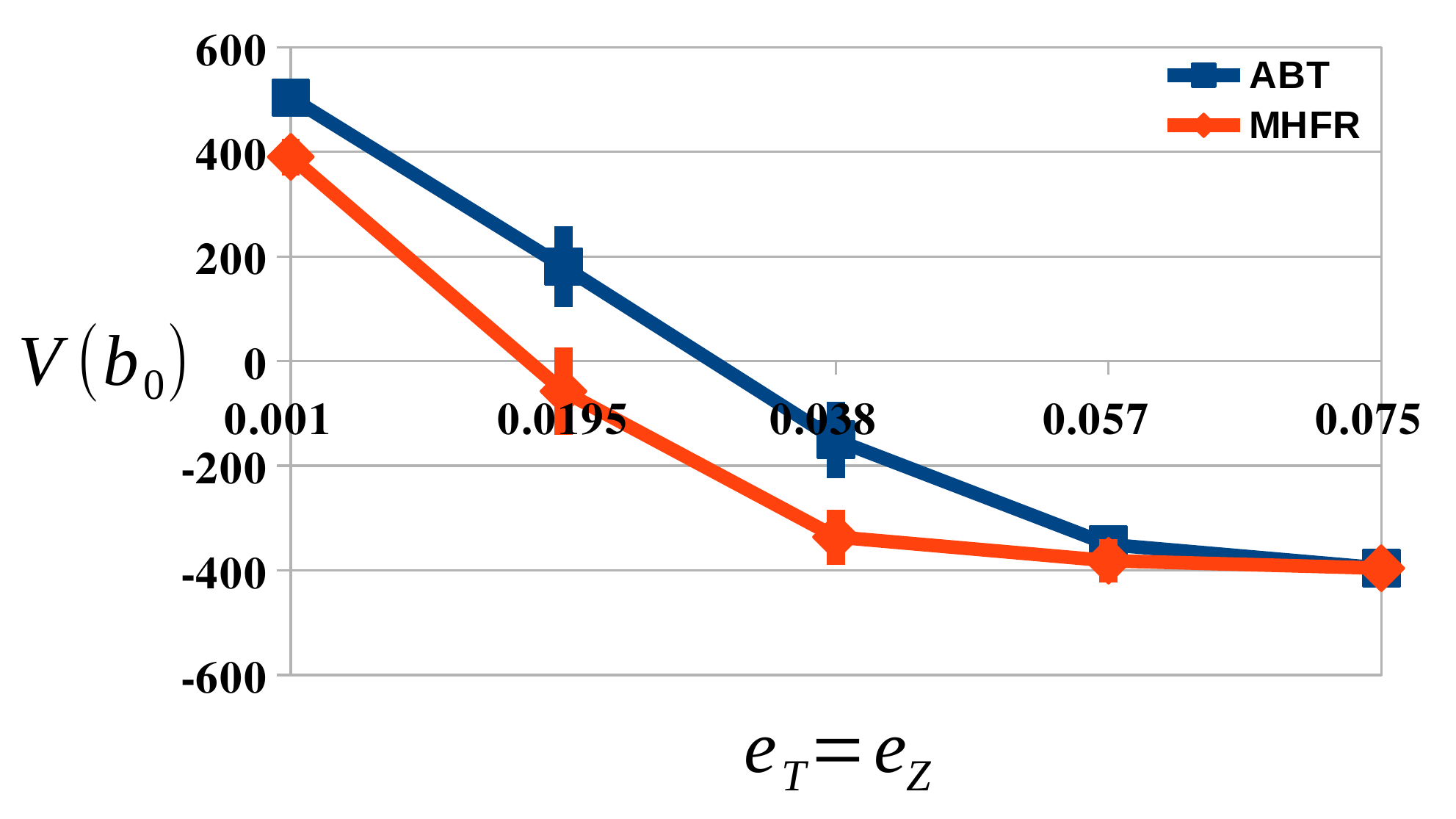}
\caption{The average total discounted rewards achieved by ABT and MHFR in the Maze scenario, as the uncertainties increase. Vertical bars are the 95\% confidence intervals.}
\label{f:relValCarMaze}
\end{figure}

An interesting remark regarding the results for the Maze scenario in \tref{t:measureCompareClutter}(b) is that the relative value difference actually decreases for large uncertainties. The reason for this can be seen in \fref{f:relValCarMaze}. As the uncertainties increase, the problem becomes so difficult, such that both solvers fail to compute a reasonable policy within the given planning time. However, clearly MHFR suffers earlier from these large uncertainties compared to ABT.

\begin{table}
\centering
\resizebox{\columnwidth}{!}{%
\begin{tabular}{|c|c|c|c|}
\hline
\multicolumn{4}{|c|}{\textbf{(a) Empty environment 4DOFs-manipulator}} \\ \hline \hline
\textbf{$\pet = \pez$} & \textbf{\monNameAbbr} & \textbf{\monGAbbr} & $\mathbf{\left |\frac{V_{ABT}(b_0) - V_{MHFR}(b_0)}{V_{ABT}(b_0)}\right |}$ \\ \hline
0.001 & 0.207 & 0.548 & 0.0110 \\ \hline
0.0195 & 0.213 & 0.557 & 0.0346 \\ \hline
0.038 & 0.243 & 0.603 & 0.0385 \\ \hline
0.057 & 0.254 & 0.617 & 0.0437 \\ \hline
0.075 & 0.313 & 0.686 & 0.0470 \\ \hline \hline
\multicolumn{4}{|c|}{\textbf{(b) Empty environment Car-like robot}} \\ \hline \hline
\textbf{$\pet = \pez$} & \textbf{\monNameAbbr} & \textbf{\monGAbbr} & $\mathbf{\left |\frac{V_{ABT}(b_0) - V_{MHFR}(b_0)}{V_{ABT}(b_0)}\right |}$ \\ \hline
0.001 & 0.169 & 0.473 & 0.1426 \\ \hline
0.0195 & 0.213 & 0.479 & 0.1793 \\ \hline
0.038 & 0.295 & 0.458 & 0.1747 \\ \hline
0.057 & 0.350 & 0.476 & 0.1839 \\ \hline
0.075 & 0.395 & 0.446 & 0.2641 \\ \hline
\end{tabular}}
\caption{Average values of SNM, MonG and the relative value difference between ABT and MHFR for the 4DOFs-manipulator (a) and the car-like robot (b) operating inside empty environments.}
\label{t:measureCompareEmpty}
\end{table}

\begin{table}
\centering
\resizebox{\columnwidth}{!}{%
\begin{tabular}{|c|c|c|c|}
\hline
\multicolumn{4}{|c|}{\textbf{(a) Factory environment}} \\ \hline \hline
\textbf{$\pet = \pez$} & \textbf{\monNameAbbr} & \textbf{\monGAbbr} & $\mathbf{\left |\frac{V_{ABT}(b_0) - V_{MHFR}(b_0)}{V_{ABT}(b_0)}\right |}$ \\ \hline
0.001 & 0.293 & 0.539 & 0.0892 \\ \hline
0.0195 & 0.351 & 0.567 & 0.1801 \\ \hline
0.038 & 0.470 & 0.621 & 0.5818 \\ \hline
0.057 & 0.502 & 0.637 & 0.7161 \\ \hline
0.075 & 0.602 & 0.641 & 1.4286 \\ \hline \hline
\multicolumn{4}{|c|}{\textbf{(b) Maze environment}} \\ \hline \hline
\textbf{$\pet = \pez$} & \textbf{\monNameAbbr} & \textbf{\monGAbbr} & $\mathbf{\left |\frac{V_{ABT}(b_0) - V_{MHFR}(b_0)}{V_{ABT}(b_0)}\right |}$ \\ \hline
0.001 & 0.215 & 0.482 & 0.2293 \\ \hline
0.0195 & 0.343 & 0.483 & 1.4473 \\ \hline
0.038 & 0.470 & 0.491 & 1.1686 \\ \hline
0.057 & 0.481 & 0.497 & 0.0985 \\ \hline
0.075 & 0.555 & 0.502 & 0.0040 \\ \hline
\end{tabular}}
\caption{Average values of SNM, MonG and the relative value difference between ABT and MHFR for the 4DOFs-manipulator operating inside the Factory environment (a) and the car-like robot operating inside the Maze environment (b).}
\label{t:measureCompareClutter}
\end{table}

\subsubsection{Effects of increasingly cluttered environments. }
\begin{figure}
\centering
\begin{tabular}{cc}
\includegraphics[width=0.20\textwidth]{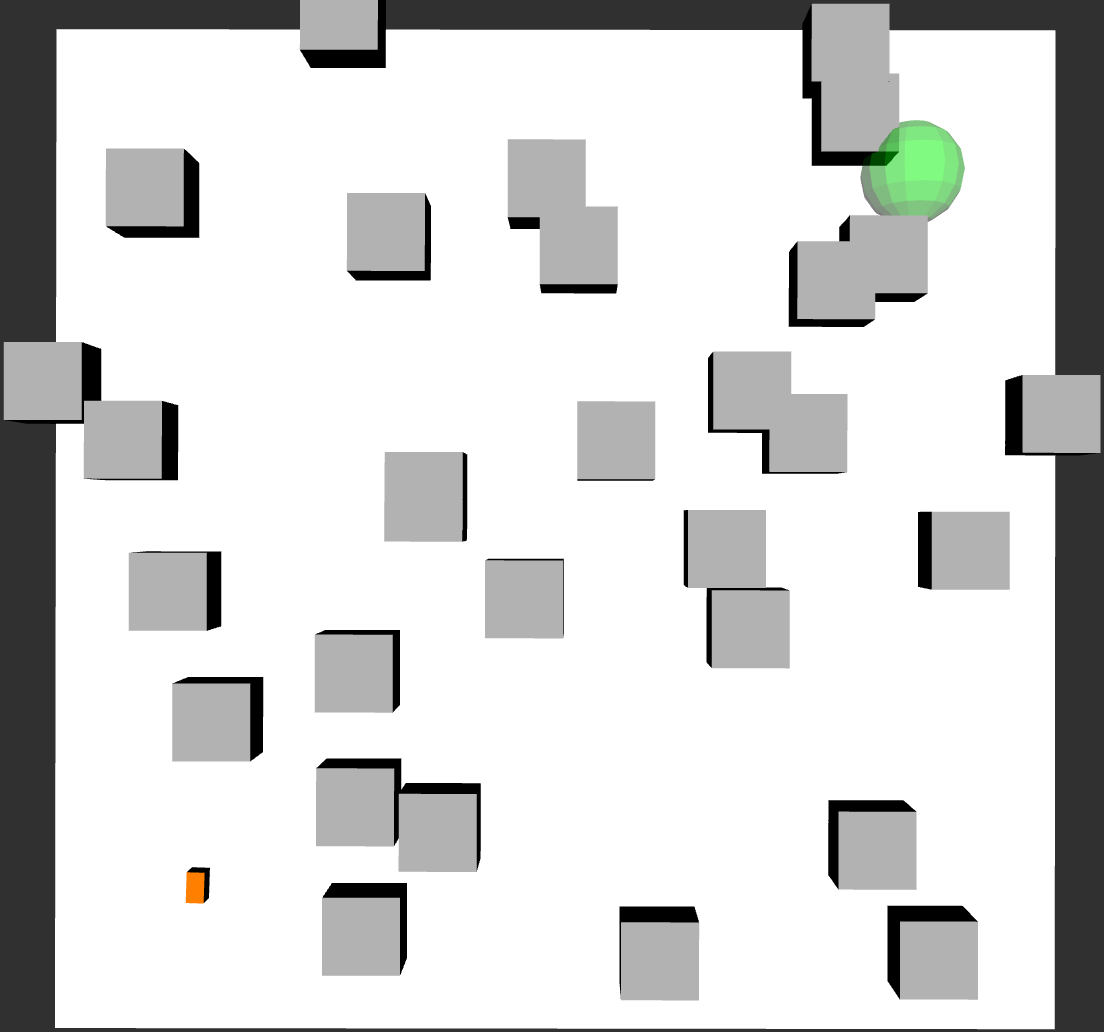} &
\includegraphics[width=0.20\textwidth]{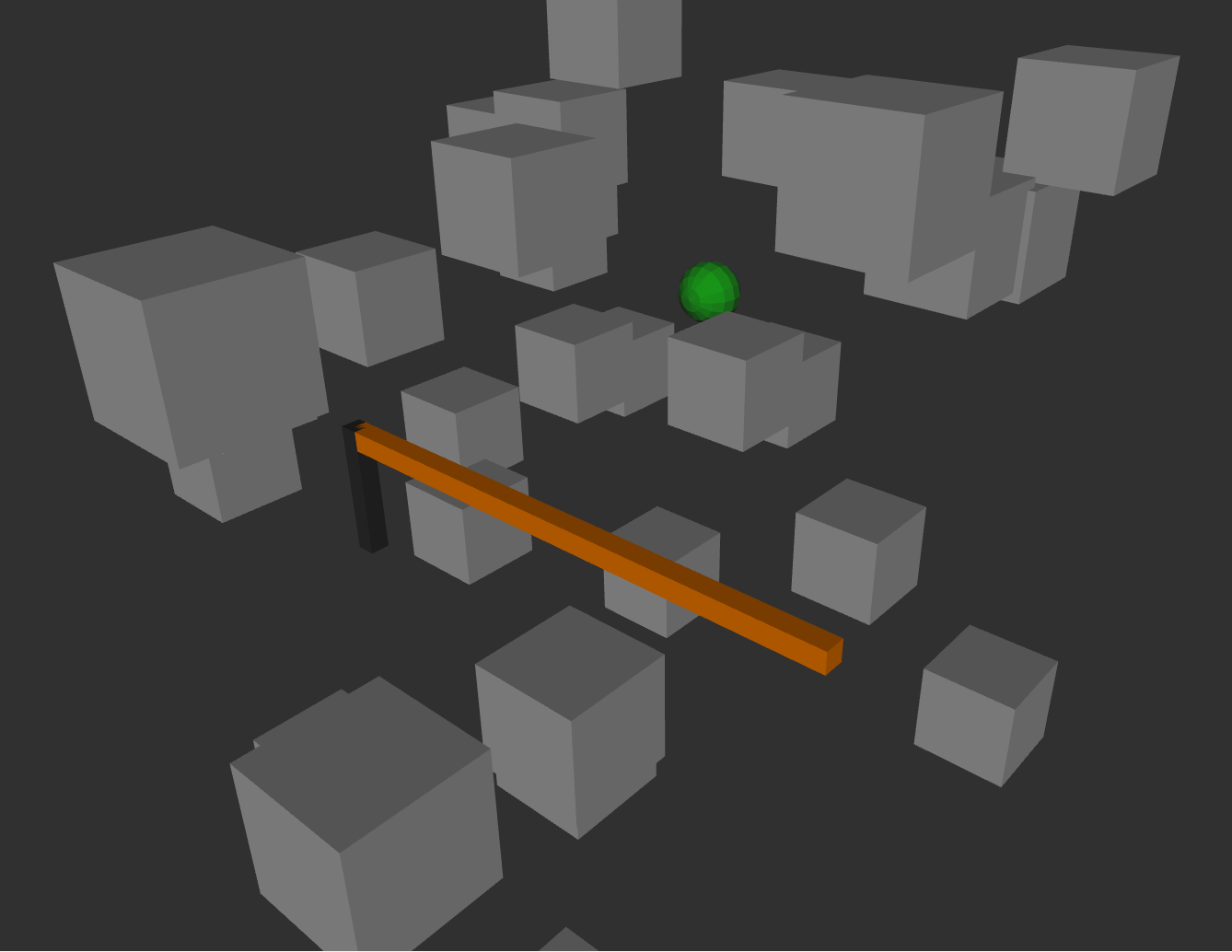} \\
(a) Car-like robot & (b) 4DOFs-manipulator
\end{tabular}
\caption{Two example scenarios for the Car-like robot (a) and the 4DOFs-manipulator (b) with 30 randomly distributed obstacles.}
\label{f:randObstacles}
\end{figure}

To investigate the effects of increasingly cluttered environments on both measures, we ran a set of experiments in which the Car-like robot and the 4DOFs-manipulator operate inside environments with an increasing number of randomly distributed obstacles. For this we generated test scenarios with 5, 10, 15, 20, 25 and 30 obstacles that are uniformly distributed across the environment. For each of these test scenarios, we randomly generated 100 environments. \fref{f:randObstacles}(a)-(b) shows two example environments with 30 obstacles for the Car-like robot and the 4DOFs-manipulator. For this set of experiments we don't take collision dynamics into account. The control and observation errors are fixed to $e_t = e_z = 0.038$ which corresponds to the median of the uncertainty values. \tref{t:increasingObstacles} presents the results for \monNameAbbr, \monGAbbr and the relative value difference between ABT ant MHFR for the 4DOFs-manipulator (a) and the car-like robot (b). From these results it is clear that, as the environments become increasingly cluttered, the advantage of ABT over MHFR increases, indicating that the obstacles have a significant effect on the Gaussian belief assumption of MHFR. Additionally \monNameAbbr is clearly more sensitive to those effects compared to \monGAbbr, whose values remain virtually unaffected by the clutterness of the environments.

\begin{table}
\centering
\resizebox{\columnwidth}{!}{%
\begin{tabular}{|c|c|c|c|}
\hline
\multicolumn{4}{|c|}{\textbf{(a) 4DOFs-manipulator with increasing number of obstacles}} \\ \hline \hline
Num obstacles & SNM & MonG & $\mathbf{\left |\frac{V_{ABT}(b_0) - V_{MHFR}(b_0)}{V_{ABT}(b_0)}\right |}$ \\ \hline 
5 & 0.359 & 0.650 & 0.0276 \\ \hline
10 & 0.449 & 0.643 & 0.0683 \\ \hline
15 & 0.514 & 0.673 & 0.2163 \\ \hline
20 & 0.527 & 0.683 & 0.2272 \\ \hline
25 & 0.651 & 0.690 & 0.2675 \\ \hline
30 & 0.698 & 0.672 & 0.3108 \\ \hline \hline
\multicolumn{4}{|c|}{\textbf{(b) Car-like robot with increasing number of obstacles}} \\ \hline \hline
Num obstacles & SNM & MonG & $\mathbf{\left |\frac{V_{ABT}(b_0) - V_{MHFR}(b_0)}{V_{ABT}(b_0)}\right |}$ \\ \hline 
5 & 0.327 & 0.459 & 0.0826 \\ \hline
10 & 0.387 & 0.473 & 0.1602 \\ \hline
15 & 0.446 & 0.482 & 0.1846 \\ \hline
20 & 0.468 & 0.494 & 0.4813 \\ \hline
25 & 0.529 & 0.489 & 0.5788 \\ \hline
30 & 0.685 & 0.508 & 0.7884 \\ \hline
\end{tabular}}
\caption{Average values of SNM, MonG and relative value difference between ABT and MHFR for the 4DOFs-manipulator (a) and the car-like robot (b) operating inside environments with increasing numbers of obstacles.}
\label{t:increasingObstacles}
\end{table}

\subsubsection{Effects of collision dynamics. }
Intuitively, collision dynamics are highly non-linear effects. Here we investigate \monNameAbbr's capability in capturing these effects compared to \monGAbbr. For this, the robots are allowed to collide with the obstacles. In other words, colliding states are not terminal and the dynamic effects of collisions are reflected in the transition model. 
For the 4DOFs-manipulator these collisions are modeled as additional constraints (contact points) that are resolved by applying "correcting velocities" to the colliding bodies in the opposite direction of the contact normals.

For the Car-like robot, we modify the transition model \eref{eq:carDynamics} to consider collision dynamics such that
\begin{equation}\label{eq:car_coll_transition}
s_{t+1} = \begin{cases}
f_{col}(s_t, a_t, v_t) & \text{ if } f(s_t, a_t, v_t)\ \text{collides} \\ 
f(s_t, a_t, v_t) & \text{ else }
\end{cases}
\end{equation}
where 
\begin{equation}\label{eq:car_coll_funct}
f_{coll}(s_t, a_t, v_t) = \left [x_t, y_t, \theta_t, -3v_t \right ]^T
\end{equation}

This transition function causes the robot to slightly "bounce" off obstacles upon collision. There are two interesting remarks regarding this transition function: The first one is that \eref{eq:car_coll_funct} is a deterministic. In other words, a collision causes an immediate reduction of the uncertainty regarding the state of the robot. Second, while the collision effects \eref{eq:car_coll_funct} are linear, \eref{eq:car_coll_transition} is not smooth since the collision dynamics induce discontinuities when the robot operates in the vicinity of obstacles.

\begin{table}
\centering
\resizebox{\columnwidth}{!}{%
\begin{tabular}{|c|c|c|c|}
\hline
\multicolumn{4}{|c|}{\textbf{(a) Maze environment with collision dynamics}} \\ \hline \hline
\textbf{$\pet = \pez$} & \textbf{\monNameAbbr} & \textbf{\monGAbbr} & $\mathbf{\left |\frac{V_{ABT}(b_0) - V_{MHFR}(b_0)}{V_{ABT}(b_0)}\right |}$ \\ \hline
0.001 & 0.425 & 0.490 & 0.3807 \\ \hline
0.0195 & 0.576 & 0.505 & 7.0765 \\ \hline
0.038 & 0.636 & 0.542 & 8.6847 \\ \hline
0.057 & 0.740 & 0.569 & 2.0194 \\ \hline
0.075 & 0.776 & 0.611 & 1.7971 \\ \hline \hline
\multicolumn{4}{|c|}{\textbf{(b) Factory environment with collision dynamics}} \\ \hline \hline
\textbf{$\pet = \pez$} & \textbf{\monNameAbbr} & \textbf{\monGAbbr} & $\mathbf{\left |\frac{V_{ABT}(b_0) - V_{MHFR}(b_0)}{V_{ABT}(b_0)}\right |}$ \\ \hline
0.001 & 0.492 & 0.639 & 0.07141 \\ \hline
0.0195 & 0.621 & 0.621 & 0.4007 \\ \hline
0.038 & 0.725 & 0.738 & 0.6699 \\ \hline
0.057 & 0.829 & 0.742 & 1.0990 \\ \hline
0.075 & 0.889 & 0.798 & 1.7100 \\ \hline
\end{tabular}}
\caption{Average values of SNM, MonG and the relative value difference between ABT and MHFR for the 4DOFs-manipulator operating inside the Factory environment (a) and the car-like robot operating inside the Maze environment (b) while being subject to collision dynamics.}
\label{t:measureCompareColl}
\end{table}

\tref{t:measureCompareColl} shows the comparison between \monNameAbbr and \monGAbbr and the relative value difference between ABT and MHFR for the 4DOFs-manipulator operating inside the Factory environment (a) and the car-like robot operating inside the Maze environment (b) while being subject to collision dynamics. It can be seen that the additional non-linear effects are captured well by \monNameAbbr. Interestingly, compared to the results in \tref{t:measureCompareClutter}(a), where the 4DOFs-manipulator operates in the same environment without collision dynamics, \monGAbbr captures the effects of collision dynamics as well, which indicates that collision dynamics have a large effect on the Gaussian assumption made by MHFR. Looking at the relative value difference between ABT and MHFR confirms this. MHFR suffers more from the increased non-linearity of the problems caused by collision dynamics compared to ABT. This effect aggravates as the uncertainty increases, which is a clear indication that the problem becomes increasingly non-linear with larger uncertainties.
Looking at the results for the car-like robot operating in the Maze scenario presents a similar picture. Comparing the results in \tref{t:measureCompareColl}(b) where collision dynamics are taken into account to \tref{t:measureCompareClutter}(b), shows that collision dynamics have a significant effect both to \monNameAbbr as well as \monG.

\subsubsection{Effects of non-linear observation functions with non-additive errors. }\label{ss:observationComparison}
In the previous experiments we assumed that the observation functions are non-linear functions with additive Gaussian noise, a special class of non-linear observation functions. This class of observation functions has some interesting implications: First of all, the resulting observation distribution remains Gaussian. This in turn means that \monGAbbr for the observation function evaluates to zero. Second, linearizing the observation function results in a Gaussian distribution with the same mean but different covariance. We therefore expect that the observation component \monNameAbbr remains small, even for large uncertainties.     
To investigate how \monNameAbbr reacts to non-linear observation functions with non-additive noise, we ran a set of experiments for the 4DOFs-manipulator operating inside the Factory environment and the car-like robot operating inside the Maze environment where we replaced  both observation functions with non-linear functions with non-additive noise.
For the 4DOFs-manipulator we replaced the observation function defined in \eref{eq:obs4DOF} with
\begin{equation}\label{e:4DOFObsNonAdditive}
o_t = g(s_t + w_t)
\end{equation}
where $w_t \sim N(0, \Sigma_w)$. In other words, the manipulator has only access to a sensor that measure the position of the end-effector in the workspace.

For the car-like robot we use the following observation function:
\begin{equation}\label{e:obstFunctCarMazeNonAdditive}
o_t = \begin{bmatrix}
\frac{1}{((x_t + w_t^1 - \hat{x}_1)^2 + (y_t + w_t^2 - \hat{y}_1)^2 + 1)}\\ 
\frac{1}{((x_t + w_t^1 - \hat{x}_2)^2 + (y_t + w_t^2 - \hat{y}_2)^2 + 1)}\\ 
v_t + w_t^3
\end{bmatrix}
\end{equation}

where $\left (w_t^1, w_t^2, w_t^3 \right )^T \sim N(0, \Sigma_w)$.
For both robots, we set $e_t = 0.038$.

\begin{table}
\centering
\resizebox{\columnwidth}{!}{%
\begin{tabular}{|c|c|c|c|c|c|}
\hline
\multicolumn{6}{|c|}{\textbf{(a) Factory environment with additive observation errors}} \\ \hline \hline
$\mathbf{\pez}$ & 0.001 & 0.0195 & 0.038 & 0.057 & 0.075 \\ \hline
\textbf{\monNameAbbr} & 0.001 & 0.004 & 0.013 & 0.036 & 0.047 \\ \hline
\textbf{MonG} & 0.0 & 0.0 & 0.0 & 0.0 & 0.0 \\ \hline \hline
\multicolumn{6}{|c|}{\textbf{(b) Factory environment with non-additive observation errors}} \\ \hline \hline
$\mathbf{\pez}$ & 0.001 & 0.0195 & 0.038 & 0.057 & 0.075 \\ \hline
\textbf{\monNameAbbr} & 0.012 & 0.087 & 0.173 & 0.234 & 0.317 \\ \hline
\textbf{MonG} & 0.0 & 0.047 & 0.094 & 0.136 & 0.173 \\ \hline
\end{tabular}}
\caption{Comparison between the observation component of \monNameAbbr and \monGAbbr for the 4DOF-manipulator operating inside the Factory environment with observation function \eref{eq:obs4DOF} (a) and \eref{e:4DOFObsNonAdditive} (b) as the observation errors increase.}
\label{t:compAdditive4DOF}
\end{table}

\begin{table}
\centering
\resizebox{\columnwidth}{!}{%
\begin{tabular}{|c|c|c|c|c|c|}
\hline
\multicolumn{6}{|c|}{\textbf{(a) Maze environment with additive observation errors}} \\ \hline \hline
$\mathbf{\pez}$ & 0.001 & 0.0195 & 0.038 & 0.057 & 0.075 \\ \hline
\textbf{\monNameAbbr} & 0.002 & 0.012 & 0.037 & 0.048 & 0.060 \\ \hline
\textbf{MonG} & 0.0 & 0.0 & 0.0 & 0.0 & 0.0 \\ \hline \hline
\multicolumn{6}{|c|}{\textbf{(b) Maze environment with non-additive observation errors}} \\ \hline \hline
$\mathbf{\pez}$ & 0.001 & 0.0195 & 0.038 & 0.057 & 0.075 \\ \hline
\textbf{\monNameAbbr} & 0.083 & 0.086 & 0.101 & 0.198 & 0.207 \\ \hline
\textbf{MonG} & 0.0 & 0.012 & 0.032 & 0.053 & 0.075 \\ \hline
\end{tabular}}
\caption{Comparison between the observation component of \monNameAbbr and \monGAbbr for the car-like robot operating inside the Maze environment with observation function \eref{e:obstFunctCarMazeAdditive}(a) and observation function \eref{e:obstFunctCarMazeNonAdditive}(b) as the observation errors increase.}
\label{t:compAdditiveCar}
\end{table}

\tref{t:compAdditive4DOF} shows the values for the observation components of \monNameAbbr and \mongAbbr for the 4DOFs-manipulator operating inside the Factory environment as the observation errors increase. As expected, for additive Gaussian errors, \monGAbbr is zero, whereas \monNameAbbr is small but measurable. This shows that \monNameAbbr is able to capture the difference of the variance between the original and linearized observation functions. For non-additive errors, the observation function is non-Gaussian, therefore we can see that both measures increase as the observation errors increase. Interestingly for both measures the observation components yield significantly smaller values compared to the transition components. This indicates that the non-linearity of the problem stems mostly from the transition function. 
For the car-like robot operating inside the Maze environment we see a similar picture. For the observation function with additive Gaussian errors, \tref{t:compAdditiveCar}(a) shows that \monGAbbr remains zero for all values of \pez, whereas \monNameAbbr yields a small but measurable value. Again, both measures increase significantly in the non-additive error case in \tref{t:compAdditiveCar}(b).

The question is now, how do ABT and MHFR perform in both scenarios when observation functions with non-additive Gaussian errors are used? \tref{t:measureCompareNonAdditive}(a) shows this relative value difference for the 4DOFs-manipulator operating inside the Factory environment. It can be seen that as the errors increase, the relative value difference between ABT and MHFR increase significantly, compared to the relative value difference shown in \tref{t:measureCompareClutter}(a), where an observation function with additive errors is used. Similarly, for the car-like robot operating inside the Maze scenario using the observation function with non-additive errors, the relative value difference shown in table \tref{t:measureCompareNonAdditive}(b) between the two solvers is much larger compared to \tref{t:measureCompareClutter}(b).

This is in line with our intuition that non-Gaussian observation functions are more challenging for linearization-based solvers.

\begin{table}
\centering
\resizebox{\columnwidth}{!}{%
\begin{tabular}{|c|c|c|c|}
\hline
\multicolumn{4}{|c|}{\textbf{(a) Factory environment with non-additive observation errors}} \\ \hline \hline
\textbf{$\pet = \pez$} & \textbf{\monNameAbbr} & \textbf{\monGAbbr} & $\mathbf{\left |\frac{V_{ABT}(b_0) - V_{MHFR}(b_0)}{V_{ABT}(b_0)}\right |}$ \\ \hline
0.001 & 0.012 & 0.0 & 0.06992 \\ \hline
0.0195 & 0.0878 & 0.0476 & 0.43861 \\ \hline
0.038 & 0.1732 & 0.0941 & 0.89720 \\ \hline
0.057 & 0.2347 & 0.1363 & 1.46063 \\ \hline
0.075 & 0.3178 & 0.1740 & 8.34832 \\ \hline \hline
\multicolumn{4}{|c|}{\textbf{(b) Maze environment with non-additive observation errors}} \\ \hline \hline
\textbf{$\pet = \pez$} & \textbf{\monNameAbbr} & \textbf{\monGAbbr} & $\mathbf{\left |\frac{V_{ABT}(b_0) - V_{MHFR}(b_0)}{V_{ABT}(b_0)}\right |}$ \\ \hline
0.001 & 0.0837 & 0.0 & -0.12451 \\ \hline
0.0195 & 0.0868 & 0.0121 & 0.33872 \\ \hline
0.038 & 0.1017 & 0.0321 & 1.41429 \\ \hline
0.057 & 0.1983 & 0.0531 & 8.70111 \\ \hline
0.075 & 0.2072 & 0.0758 & 0.95132 \\ \hline
\end{tabular}}
\caption{Average values of SNM, MonG and the relative value difference between ABT and MHFR for the 4DOFs-manipulator operating inside the Factory environment (a) and the car-like robot operating inside the Maze environment (b) with non-additive observation errors}
\label{t:measureCompareNonAdditive}
\end{table}

\subsection{Testing \nop}\label{ssec:testing_snm-planner}
In this set of experiments we want to test the performance of \nop in comparison with the two component planners ABT and MHFR. To this end we tested \nop on three problem scenarios: The Maze scenario for the car like robot shown in \fref{f:scenarioCompStudy}(a) and the Factory scenario for the 4DOFs-manipulator. Additionally we tested \nop on a scenario in which the Kuka iiwa robot operates inside an office environment, as shown in \fref{f:scenarioCompStudy}(b). Similarly to the Factory scenario, the robot has to reach a goal area while avoiding collisions with the obstacles. The planning time per step is 8s in this scenario. For the \monNameAbbr-threshold we chose 0.5. Here we set $\pet=\pez=0.038$.

\begin{table}
\centering
\resizebox{\columnwidth}{!}{%
\begin{tabular}{|c|c|c|c|}
\hline
\textbf{Planner} & \textbf{Car-like robot} & \textbf{4DOFs-manipulator} & \textbf{Kuka iiwa} \\ \hline \hline
ABT & -150.54 $\pm$ 40.6 & 801.78 $\pm$ 25.7 & 498.33 $\pm$ 30.6 \\ \hline
MHFR & -314.25 $\pm$ 31.4 & 345.82 $\pm$ 60.8 & -163.21 $\pm$ 29.6 \\ \hline
\nop & \textbf{14.68 $\pm$ 46.3} & \textbf{833.17 $\pm$ 13.4} & \textbf{620.67 $\pm$ 35.7} \\ \hline
\end{tabular}}
\caption{Average total discounted reward and $\pm$ 95\% confidence interval over 1,000 simulation runs. The proportion of ABT being used in the Maze, Factory and Office scenarios is 37.85\%, 56.43\% and 42.33\% respectively.}
\label{t:snmPlannerResults}
\end{table}

\begin{figure}
\centering
\includegraphics[width=0.5\columnwidth]{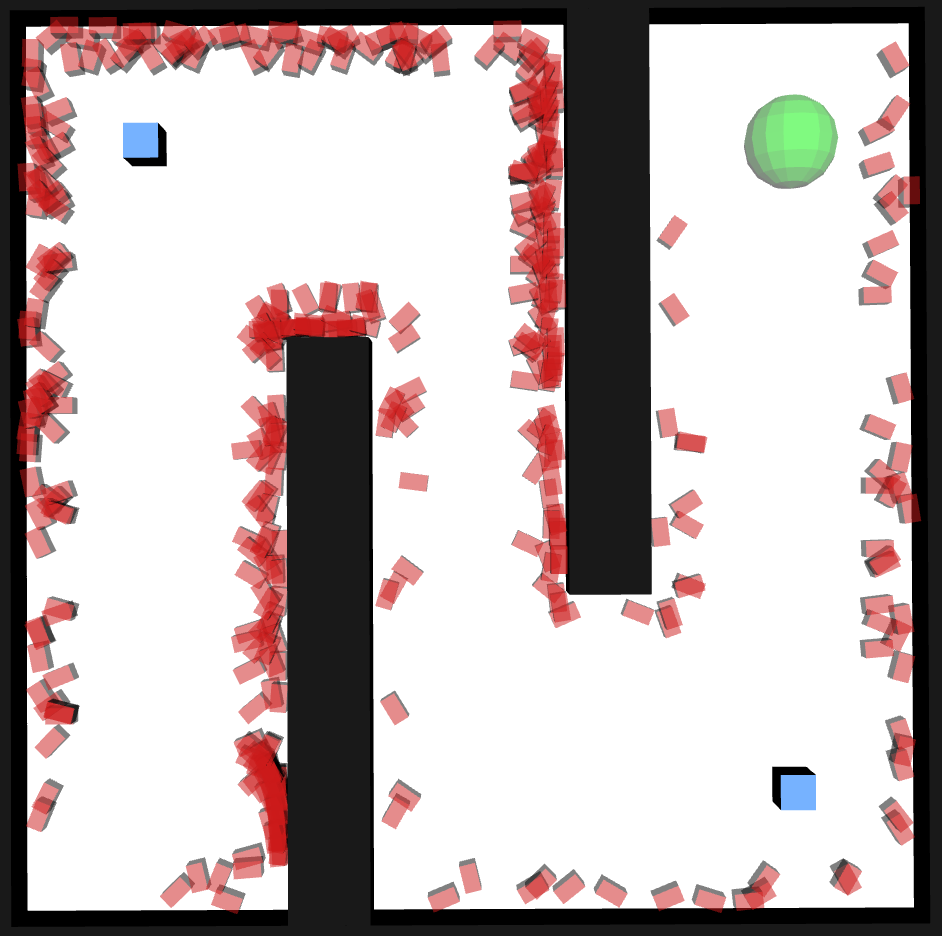}
\caption{State samples in the Maze scenario for which the approximated \monNameAbbr value exceeds the chosen threshold of 0.5}
\label{f:DubinSNMSamples}
\end{figure}

The results in \tref{t:snmPlannerResults} indicate that \nop is able to approximately identify when it is beneficial to use a linearization-based solver and when a general solver should be used. In all three scenarios, \nop outperforms the two component planners. In the Maze scenario, the difference between \nop and the component planners is significant. The reason is, MHFR is well suited to compute a long-term strategy, as it constructs nominal trajectories from the current state estimate all the way to the goal, whereas the planning horizon of ABT is limited by the depth of the search tree. However, in the proximity of obstacles, the Gaussian belief assumption of MHFR are no long valid, and careful planning is required to avoid collisions with the obstacles. In general ABT handles these situations better than MHFR. \nop combines the benefits of both planners and alleviates their shortcoming. \fref{f:DubinSNMSamples} shows state samples for which the \monNameAbbr-values exceed the given threshold of 0.5. It is obvious that many of these samples are clustered around obstacles. In other words, when the support set of the current belief (i.e. the subset of the state space that is covered by the belief particles) lies in open areas, MHFR is used to drive the robot towards the goal, whereas in the proximity of obstacles, ABT is used to compute a strategy that avoids collisions with the obstacles.

A similar behavior was observed in the KukaOffice environment. During the early planning steps, when the robot operates in the open area, MHFR is well suited to drive the end-effector towards the goal area, but near the narrow passage at the back of the table, ABT in general computes better motion strategies. Again, \nop combines both strategies to compute better motion strategies than each of the component planners alone.

\subsubsection{Sensitivity of \nop. }
In this experiment we test how sensitive the performance of \nop is to the choice of the \monNameAbbr-threshold. Recall that \nop uses this threshold to decide, based on a local approximation of \monNameAbbr, which solver to use for the policy computation. For small thresholds \nop favors ABT, whereas for large thresholds MHFR is favored.

For this experiment we test \nop on the Factory problem (\fref{f:scenarioCompStudy}(b)) with multiple values for the \monNameAbbr-threshold, ranging from 0.1 to 0.9. For each threshold value we estimate the average total expected discounted reward achieved by \nop using 1,000 simulation runs. Here we set $\pet=\pez=0.038$. 

\tref{t:SNMPlannerSensitivity} summarizes the results. It can be seen that the choice of the threshold can affect the performance of \nop, particularly for values that are on either side of the spectrum (very small values or very large values) where \nop favors only one of the component solvers. However, between the threshold values of 0.2 and 0.5 the results are fairly consistent, which indicates that there's a range of \monNameAbbr-threshold values for which \nop performs well.
\begin{table}[htb]
\centering
\resizebox{\columnwidth}{!}{%
\begin{tabular}{|c|c|c|}
\hline
\textbf{SNM-Threshold} & \textbf{Avg. total discounted reward} & \textbf{\% ABT used}\\
\hline
\hline
0.1 & 789.43 $\pm$ 18.4 & 100.0 \\
\hline
0.2 & 794.69 $\pm$ 15.3 & 95.3 \\
\hline
0.3 & 801.82 $\pm$ 14.2 & 89.8 \\
\hline
0.4 & 834.32 $\pm$ 13.3 & 65.2\\
\hline
0.5 & 833.17 $\pm$ 13.4 & 59.6\\
\hline
0.6 & 725.71 $\pm$ 19.6 & 42.7 \\
\hline
0.7 & 622.39 $\pm$ 18.5 & 30.6 \\
\hline
0.8 & 561.02 $\pm$ 29.4 & 21.5 \\
\hline
0.9 & 401.79 $\pm$ 39.6 & 7.8 \\
\hline
\end{tabular}}
\caption{Average total discounted reward and 95\% confidence intervals of \nop on the Factory problem for varying \monNameAbbr-threshold values. The average is collected over 1,000 simulation runs. The last column shows the percentage of ABT being used as the component solver.\vspace{-9pt}}
\label{t:SNMPlannerSensitivity}
\end{table}
\vspace{-9pt}

\section{Summary and Future Work}\label{sec:discussion}
This paper presents our preliminary work in identifying the suitability of linearization for motion planning under uncertainty. To this end, we present a general measure of non-linearity, called \monName (\monNameAbbr), which is based on the distance between the distributions that represent the system's motion--sensing model and its linearized version. Comparison studies with one of state-of-the-art methods for non-linearity measure indicate that \monNameAbbr is more suitable in taking into account obstacles in measuring the effectiveness of linearization.

We also propose a simple on-line planner that uses a local estimate of \monNameAbbr  to select whether to use a general POMDP solver or a linearization-based solver for robot motion planning under uncertainty. Experimental results indicate that our simple planner can appropriately decide where linearization should be used and generates motion strategies that are comparable or better than each of the component planner.

Future work abounds. For instance, the question for a better measure remains. The total variation distance relies on computing a maximization, which is often difficult to estimate. Statistical distance functions that relies on expectations exists and can be computed faster. How suitable are these functions as a non-linearity measure? Furthermore, our upper bound result is relatively loose and can only be applied as a sufficient condition to identify if linearization will perform well. It would be useful to find a tighter bound that remains general enough for the various linearization and distribution approximation methods in robotics.

\section{Acknowledgements}
This work is partially funded by ANU Futures Scheme QCE20102. The early part of this work is funded by UQ and CSIRO scholarship for Marcus Hoerger. 

\bibliographystyle{ieeetr}
\bibliography{References}

\newpage
\appendix
\section{Appendix}\label{s:appendix}
For writing compactness we use the following shorthand notations for the transition and observation functions thoughout the next two subsections: $\trans = \transComp$, $\linTrans = \linTransComp$ and $\obsF = \obsFComp$, $\linObsF = \linObsFComp$. Additionally in section \sref{ssec:proofLem4} we use the notations $\trans_{k} = T(\st_k, \act, \stp)$ and $\linTrans_{k} = \widehat{T}(\st_k, \act, \stp)$.
\subsection{Proof of Lemma \ref{lem:lem0}}\label{ssec:lemma_0_proof}
Consider any $\sigma \in \Gamma$ with its action $\act \in \actSpace$ and observation strategy $\nu$. Then for any $\st\in\stSpace$ 
\begin{align}\label{eq:first_lemma_0}
    &&&\left | \alpha_{\sigma}(\st) - \widehat{\alpha}_{\sigma}(\st) \right | \nonumber \\
    &&=&\left | \rewFuncComp{\st}{\act} + \gamma \intSDash \intO \trans \obsF \alpha_{\nu(\obs)}(\stp)\de\obs \de\stp \right. \nonumber\\ 
    &&&\left. -\rewFuncComp{\st}{\act} - \gamma \intSDash \intO \linTrans \linObsF \widehat{\alpha}_{\nu(\obs)}(\stp) \de\obs \de\stp \right | \nonumber \\
    &&=&\gamma \left | \intSDash \intO \trans \obsF \alpha_{\nu(\obs)}(\stp) - \linTrans\linObsF \widehat{\alpha}_{\nu(\obs)}(\stp) \de\obs \de\stp \right | \nonumber \\
    &&\leq&\gamma \left ( \left |\intSDash \intO \trans \obsF \left [\alpha_{\nu(\obs)}(\stp) -\widehat{\alpha}_{\nu(\obs)}(\stp) \right ] \de\obs \de\stp \right | \right. \nonumber \\
    & && + \left. \left |\intSDash \intO \widehat{\alpha}_{\nu(\obs)}(\stp) \left [\trans\obsF -\linTrans \linObsF \right ] \de\obs \de\stp\right | \right )
\end{align}

Let's have a look at the second term on the right-hand side of \eref{eq:first_lemma_0}, that is
\begin{align}\label{eq:first_lemma_1}
&&term2(\st, \act) =& \left |   \intSDash \intO \widehat{\alpha}_{\nu(\obs)}(\stp) \left [\trans \obsF - \linTrans \linObsF \right ] \de\obs \de\stp \right |
\end{align}

We can expand this term as follows:
\begin{align}\label{eq:first_lemma_2}
&&&term2(\st, \act) \nonumber \\
&&=& \left |  \intSDash \intO \widehat{\alpha}_{\nu(\obs)}(\stp) \left [ \trans \obsF - \linTrans\obsF + \linTrans\obsF - \linTrans \linObsF\right ] \de\obs \de\stp \right |\nonumber \\
&&\leq& \left | \intSDash \left [ \trans - \linTrans\right ] \intO \widehat{\alpha}_{\nu(\obs)}(\stp)\obsF \de\obs \de\stp\right | \nonumber \\
& && +  \left |\intSDash \linTrans \intO \widehat{\alpha}_{\nu(\obs)}(\stp)\left [\obsF - \linObsF \right ] \de\obs\de\stp\right | \nonumber \\
&&\leq& \intSDash \left |\trans- \linTrans \right | \intO \left | \widehat{\alpha}_{\nu(\obs)}(\stp) \right | \obsF \de\obs\de\stp \nonumber \\
& && + \intSDash \linTrans \intO \left | \widehat{\alpha}_{\nu(\obs)}(\stp) \right | \left |\obsF - \linObsF \right | \de\obs\de\stp
\end{align}

The term $\left | \widehat{\alpha}_{\nu(\obs)}(\stp) \right |$ can be upper-bounded via $\left | \widehat{\alpha}_{\nu(\obs)}(\stp)\right | \leq \frac{R_{m}}{1-\gamma}$ for any $\st \in \stSpace$, which yields
\begin{align}\label{eq:first_lemma_3}
&&&term2(\st, \act) \nonumber \\
&&\leq& \frac{R_{m}}{1-\gamma} \left [\intSDash \left |\trans - \linTrans \right | \de\stp + \intSDash\linTrans\intO\left | \obsF - \linObsF \right | \de\obs\de\stp \right ]
\end{align}

From the definition of the total variation distance, it follows that $\intSDash \left | \trans - \linTrans\right |\de\stp = 2D_{TV}^{\st, \act}(\trans, \linTrans)$ for any given $\st \in \stSpace$ and $\act \in \actSpace$ and $\intO \left |\obsF - \linObsF \right | \de\obs= 2D_{TV}^{\stp, \act}(\obsF, \linObsF)$ for any given $\stp \in \stSpace$. Substituting these equalities into \eref{eq:first_lemma_3} and taking the supremum over the conditionals $\st, \stp$ and $\act$ allows us to upper-bound \eref{eq:first_lemma_3} by
\begin{equation}
term2(\st, \act) \leq 2\frac{R_{m}}{1-\gamma} \nm{\pomdp}{\linPomdp}
\end{equation}

Substituting this upper bound into \ref{eq:first_lemma_0} yields
\begin{align}\label{eq:lemm2_eq_5}
&&&\left | \alpha_{\sigma}(\st) - \widehat{\alpha}_{\sigma}(\st) \right | \nonumber \\
&&\leq& \gamma\biggl | 2\frac{R_{m}}{1-\gamma} \nm{\pomdp}{\linPomdp} \biggr. \nonumber \\
&&&\left. +  \intSDash\intO \trans \obsF \left [\alpha_{\nu(\obs)}(\stp) - \widehat{\alpha}_{\nu(\obs)}(\stp) \right] \de\obs \de\stp \right | \nonumber \\
&&\leq& \gamma \biggl ( 2 \frac{R_{m}}{1-\gamma} \nm{\pomdp}{\linPomdp} \biggr.\nonumber \\
    &&&\left. + \intSDash\intO \trans \obsF \left |\alpha_{\nu(\obs)}(\stp) - \widehat{\alpha}_{\nu(\obs)}(\stp) \right| \de\obs \de\stp \right )
\end{align}

The last term on the right hand side of \ref{eq:lemm2_eq_5} is essentially a recursion. Unfolding this recursion yiels

\begin{equation}
\left | \alphaPiS{\st} - \alphaPiHatS{\st} \right | \leq 2\gamma\frac{R_{m}}{(1-\gamma)^2} \nm{\pomdp}{\linPomdp}
\end{equation}

which is Lemma \ref{lem:lem0} $\square$

\subsection{Proof of \lref{lem:lemApprox}}\label{ssec:proofLem4}
We can write the absolute difference between the \monNameAbbr-values conditioned on two states $\st_1, \st_2 \in \stSpace_i$ as
\begin{align}
&&&\left | \nmSymT(\st_1) - \nmSymT(\st_2) \right | \nonumber \\
&&=& \left | \sup_{\act \in \actSpace} D_{TV}(\trans_1, \linTrans_1) - \sup_{\act \in \actSpace} D_{TV}(\trans_2, \linTrans_2) \right | \nonumber \\
&&=&\left |\frac{1}{2} \sup_{\act \in \actSpace} \sup_{\left |f \right | \leq 1} \left | \intSDash f(\stp) \left [\trans_1 - \linTrans_1 \right ] \de\stp \right| \right. \nonumber \\
&&&\left. - \frac{1}{2} \sup_{\act \in \actSpace} \sup_{\left |f \right | \leq 1}\left |\intSDash f(\stp) \left [\trans_2 - \linTrans_2 \right ]\de\stp\right | \right | 
\end{align}

Manipulating the algebra allows us to write
\begin{align}\label{e:lem3_eq2}
&&&\left | \nmSymT(\st_1) - \nmSymT(\st_2) \right | \nonumber \\
&&\leq& \frac{1}{2}\sup_{\act\in\actSpace}\left | \sup_{\left |f \right | \leq 1} \left ( \intSDash f(\stp) \left [\trans_1 - \trans_2 \right ] \de\stp \right. \right.\nonumber \\
&&&\left.\left. + \intSDash f(\stp)\left [\linTrans_1 - \linTrans_2 \right] \de\stp \right ) \right |\nonumber \\
&&\leq& \frac{1}{2}\sup_{\act \in \actSpace} \left (\sup_{\left |f \right | \leq 1}\intSDash f(\stp) \left |\trans_1 - \trans_2 \right | \de\stp \right. \nonumber \\
&&&\left. + \sup_{\left |f \right | \leq 1} \intSDash f(\stp) \left |\linTrans_1 - \linTrans_2 \right |\de\stp\right ) \nonumber \\
&&\leq& \frac{1}{2}D_{S}(\st_1, \st_2)\left [C_{\trans_i} + C_{\linTrans_i} \right ]
\end{align}

For the last inequality we bound the terms $\left | \trans_1 - \trans_2 \right |$ and $\left |\linTrans_1 - \linTrans_2 \right |$ using \dref{d:partition}. Furthermore we use the fact that $\sup_{\left |f \right | \leq 1} \intSDash f(\stp) \de\stp = 1$, assuming that the state space $\stSpace$ is normalized. This concludes the proof of \lref{lem:lemApprox}. $\square$


\end{document}